\documentclass[10pt,twocolumn,letterpaper]{article}

\usepackage{cvpr}
\usepackage{times}
\usepackage{epsfig}
\usepackage{graphicx}
\usepackage{subfigure}
\usepackage{amsmath}
\usepackage{amssymb}
\usepackage{mathrsfs}
\usepackage{algorithmic}
\usepackage{caption}
\usepackage[numbers, sort]{natbib}
\usepackage[ruled,vlined]{algorithm2e}
\usepackage{array}
\usepackage{color}

\usepackage[pagebackref=true,breaklinks=true,letterpaper=true,colorlinks,bookmarks=false]{hyperref}

\cvprfinalcopy 

\ifcvprfinal\fi
\begin{document}

\title{Nonlinear Local Metric Learning for Person Re-identification}

\author{
Siyuan Huang$^1$, Jiwen Lu$^1$, Jie Zhou$^1$, Anil K. Jain$^2$\\
$^1$Department of Automation, Tsinghua University, Beijing, China\\
$^2$Dept. Computer Science and Engineering, Michigan State University, MI, USA
}

\maketitle

\begin{abstract}
Person re-identification aims at matching pedestrians observed from non-overlapping camera views. Feature descriptor and metric learning are two significant problems in person re-identification. A discriminative metric learning method should be capable of exploiting complex nonlinear transformations due to the large variations in feature space. In this paper, we propose a nonlinear local metric learning (NLML) method to improve the state-of-the-art performance of person re-identification on public datasets. Motivated by the fact that local metric learning has been introduced to handle the data which varies locally and deep neural network has presented outstanding capability in exploiting the nonlinearity of samples, we utilize the merits of both local metric learning and deep neural network to learn multiple sets of nonlinear transformations. By enforcing a margin between the distances of positive pedestrian image pairs and distances of negative pairs in the transformed feature subspace, discriminative information can be effectively exploited in the developed neural networks. Our experiments show that the proposed NLML method achieves the state-of-the-art results on the widely used VIPeR, GRID, and CUHK 01 datasets.
\end{abstract}

\section{Introduction} 	

Person re-identification aims to recognize people who have been observed from different disjoint cameras, which has become an effective tool for people association and behavior analysis in video surveillance~\cite{vezzani2013people,gong2014person}. Due to the complex variations in illumination, pose, viewpoint, occulusion and image resolution across camera views, person re-identification still remains a challenging problem in computer vision.

\begin{figure}
\begin{center}
   \includegraphics[width=0.48\textwidth]{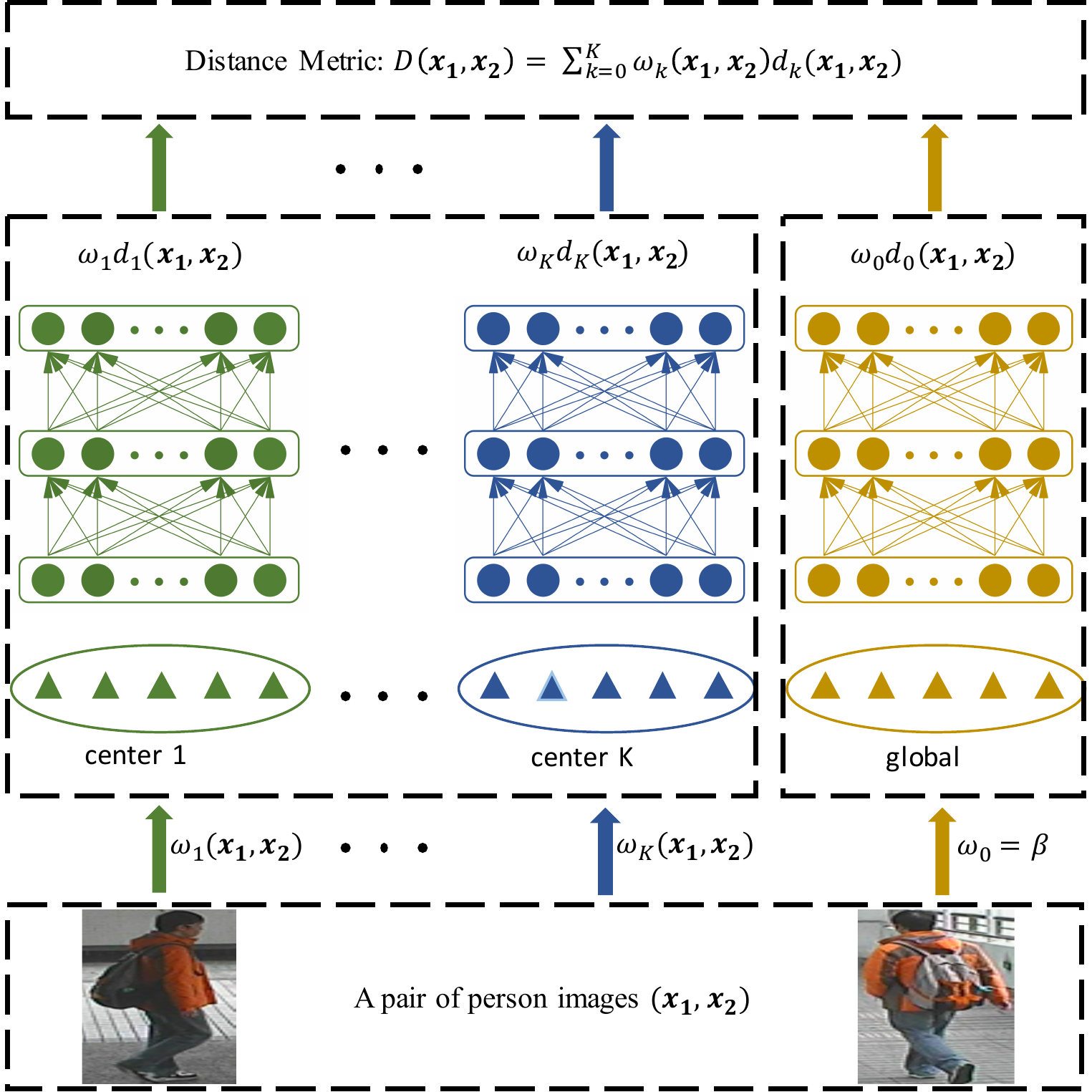}
\end{center}
   \caption{Basic idea of the proposed NLML method. The whole network consists of one global neural network and a set of local neural networks which correspond to different local clusters. For a given pair of person images $\mathbf{x}_1$ and $\mathbf{x}_2$, we compute the nonnegative weight $w_k\left(\mathbf{x}_1,\mathbf{x}_2\right)$ for $k$th local region and pass it through $K$ local and one global deep neural networks to get the representation at output layer $d_k(\mathbf{x}_1,\mathbf{x}_2)$. The final distance metric $\mathcal{D}(\mathbf{x}_1,\mathbf{x}_2)$ is defined as a linear combination of $K+1$ matrices.}
\label{fig:network}
\end{figure}

Previously proposed approaches which improve the person re-identification performance~\cite{zhao2013unsupervised, zhao2014learning, xiong2014person, li2014deepreid, liao2015person, chen2015similarity, paisitkriangkrai2015learning} can be mainly categorized into two classes: 1) developing robust descriptors to handle the variations in pedestrian images; 2) designing discriminative distance metrics to measure the similarity of pedestrian images. For the first category, different cues (color, shape, texture) from pedestrian images are employed for feature representation. Representative descriptors in person re-identification include local binary patterns (LBP)~\cite{xiong2014person}, ensemble of local feature (ELF)~\cite{gray2008viewpoint}, mid-level filter~\cite{zhao2014learning} and local maximal occurrence (LOMO)~\cite{liao2015person}. For the second category, a distance metric is learned from labeled training samples, under which the inter-class and intra-class variations of pedestrian images are increased and decreased, respectively. Typical metric learning algorithms include large margin nearest neighbor (LMNN)~\cite{weinberger2005distance}, information theoretic metric learning (ITML)~\cite{davis2007information}, and pairwise constrained component analysis (PCCA)~\cite{mignon2012pcca}.

While metric learning methods achieved good performance in many person re-identification systems~\cite{chopra2005learning, davis2007information, koestinger2012large, weinberger2005distance, xiong2014person, pedagadi2013local, li2013learning, hu2014discriminative, bohne2014large, paisitkriangkrai2015learning}, most of them learn a Mahalanobis distance metric to transform samples into a new feature space, which are not capable enough of exploiting the nonlinear relationship of pedestrian samples distributed in a nonlinear feature space due to large intra-class variations. Moreover, a single distance metric usually suffers limitations while handling data which varies locally. To address this, we propose a nonlinear local metric learning (NLML) method for person re-identification. Figure~\ref{fig:network} illustrates the basic idea of the proposed NLML method. Unlike most existing metric learning methods, NLML develops one global feed-forward neutral network and a set of local feed-forward neutral networks to jointly learn multiple sets of nonlinear transformations. The learning procedure is formulated as a large margin optimization problem and the gradient descent algorithm is employed to estimate the networks. Experimental results on the VIPeR, GRID and CUHK 01 datasets demonstrate the efficacy of the proposed NLML method.
\section{Related Work}

\textbf{Person Re-identification}: Most existing person re-identification methods can be classified into two categories: feature representation and metric learning. Feature representation methods aim to seek discriminative descriptors which are robust to variations of viewpoint, pose, and illumination in pedestrian images captured across different cameras~\cite{bazzani2012multiple, farenzena2010person, gheissari2006person, gray2008viewpoint, wang2007shape, liao2015person, zhao2014learning, ma2013domain}. Farenzena~\etal~\cite{farenzena2010person} developed a symmetry driven accumulation of local feature (SDALF) for appearance modeling of human body images. Cheng \etal~\cite{cheng2011custom} employed a pre-learned pictorial structure model to localize human body parts. Gray \etal~\cite{gray2008viewpoint} selected a subset of color and texture features for body representation which were assumed to be invariant with viewpoint change. Kviatkovsky \etal~\cite{kviatkovsky2013color} utilized invariant color descriptors to make it robust to certain illumination changes. Zhao \etal~\cite{zhao2013unsupervised} learned a distinct salience feature descriptor to distinguish the correct matched person from others. Liao \etal~\cite{liao2015person} constructed a feature descriptor which analyzes the horizontal occurrence of local features, and maximizes the occurrence to make a stable representation against viewpoint changes.

Metric learning algorithms have also been widely used in person re-identification~\cite{chopra2005learning, davis2007information, koestinger2012large, weinberger2005distance, xiong2014person, pedagadi2013local, li2013learning, hu2014discriminative, bohne2014large, paisitkriangkrai2015learning}. Compared with the distance measures such as the L1-Norm and Euclidean distance, the learned distance metrics are more discriminative to handle features which are extracted from person images across different cameras. Prosser \etal ~\cite{prosser2010person} developed a ranking model using support vector machine. Hirzer \etal ~\cite{hirzer2012person} learned a discriminative metric by using relaxed pairwise constraints. Li \etal ~\cite{li2013learning} proposed learning a Locally-Adaptive Decision Function (LADF) for person re-dientification. Loy \etal \cite{loy2013person} exploited the manifold structure of the gallery set to perform ranking. Xiong \etal ~\cite{xiong2014person} and Chen \etal ~\cite{chen2015similarity} proposed kernel based metric learning methods to exploit the nonlinearity relationship of samples in the feature space.

\textbf{Metric Learning}: Existing metric learning methods can be mainly classified into two categories: unsupervised and supervised. Unsupervised methods seek a low-dimensional subspace to preserve the geometrical information of samples. Representative unsupervised metric learning methods include principal component analysis (PCA)~\cite{wold1987principal}, locality preserving projections (LPP)~\cite{niyogi2004locality}, locally linear embedding (LLE)~\cite{roweis2000nonlinear}, and multidimensional scaling (MDS)~\cite{kruskal1978multidimensional}. Supervised methods learn a discriminative distance metric under which the intra-class variation is increased and the inter-class variation is decreased. Typical methods in this category include linear discriminant analysis (LDA)~\cite{fisher1936use}, neighborhood component analysis (NCA)~\cite{goldberger2004neighbourhood}, cosine similarity metric learning~\cite{nguyen2011cosine}, large margin nearest neighbor (LMNN)~\cite{weinberger2005distance}, and information theoretic metric learning (ITML)~\cite{davis2007information}, discriminative deep metric learning (DDML)~\cite{hu2014discriminative} and large margin local metric learning (LMLML)~\cite{bohne2014large}.

\section{Nonlinear Local Metric Learning}
\subsection{Motivations}

While metric learning techniques have achieved good performance in many person re-identification systems, they still own several drawbacks: 1) Most existing supervised metric learning methods only learn a single Mahalanobis distance metric to transform samples into a new feature space, which is not powerful enough to exploit the nonlinear relationship of samples, especially when handling heterogeneous samples; 2) While kernel-based methods can exploit the nonlinearity of samples, they usually suffer from the scalability problem; 3) Deep learning methods such as ~\cite{yi2014deep} face the problem of small training set in learning the convolutional neural network.

\begin{figure*}
\begin{center}
   \includegraphics[width=\textwidth]{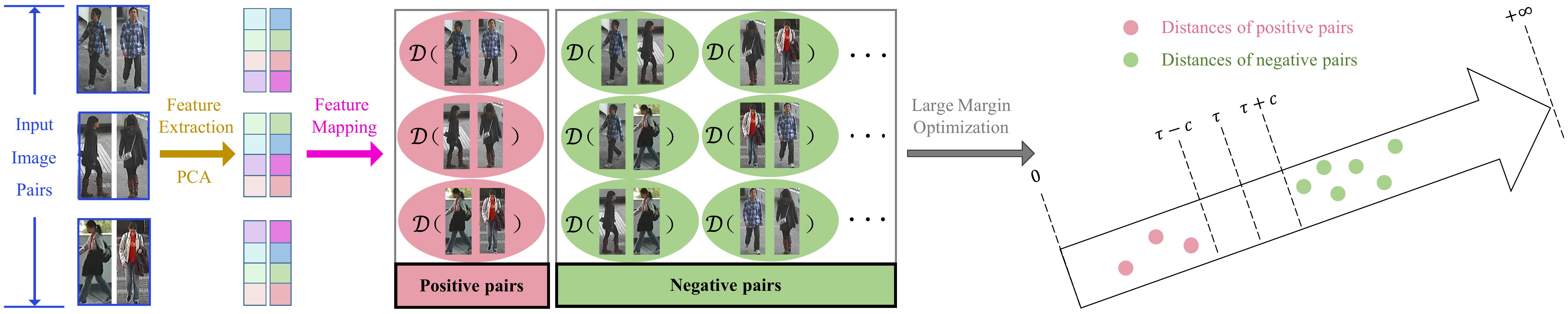}
\end{center}
   \caption{Flow chart of the proposed NLML method. For input person image pair $\mathbf{x}_i$ and $\mathbf{x}_j$, we extract the features and utilize one global and $K$ local deep neural networks to map the features and obtain the distance metric $\mathcal{D}(\mathbf{x}_i,\mathbf{x}_j)$. To exploit more discriminative information from the image representations, a large margin framework is employed to enforce the distances between positive pairs smaller than the distances between negative pairs. }
\label{fig:flowchart}
\end{figure*}
\begin{figure}
\begin{center}
   \includegraphics[width=0.48\textwidth]{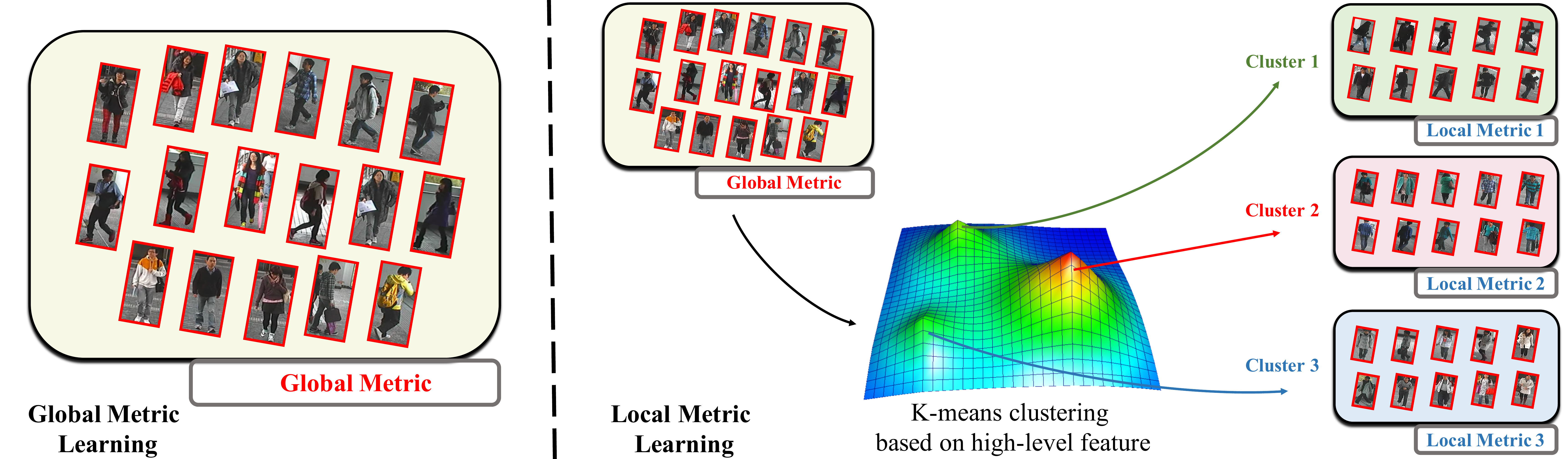}
\end{center}
   \caption{Comparison of global metric learning and local metric learning. The local metric learning combines with local metrics learning from a set of local regions.}
\label{fig:comparison}
\end{figure}

To address these limitations, we propose an effective metric learning method with a deep architecture for person re-identification. On the one hand, we model the distance metric as a combination of one global and a set of local metrics which exploit more discriminative information from the training set, comparison of local metric learning and global metric is shown in Figure~\ref{fig:comparison}. On the other hand, we replace the conventional Mahalanobis metrics with deep neural networks which can learn nonlinear similarity measures. Figure~\ref{fig:flowchart} shows the flow chart of the proposed NLML method.

\subsection{Model Architecture}

Let $ \rm{X} = [x_1,x_2,\cdots,x_N] \in \mathbb{R}^{d \times N} $ be the training set, which contains $N$ samples, where $x_i \in \mathbb{R}^d$ is a $d$-dimensional feature vector. As shown in Figure~\ref{fig:network}, we construct $K$ local deep neural networks for the corresponding $K$ local regions and one global deep neural network for the entire input space. Assume there are $M_k+1$ layers in $k$th network, and $p_k^{(m)}$ units in the $m_k$th layer, where $m_k = 1,2,\cdots, M_k$ and $k = 0,1, \cdots ,K$. For a given person image $\mathbf{x}\in\mathbb{R}^d$, the output of the first layer in the $k$th network is computed as: $\mathbf{h}_k^{(1)} = \phi(\mathbf{W}_k^{(1)}\mathbf{x}+\mathbf{b}_k^{(1)})$, where $\mathbf{W}_k^{(1)}\in\mathbb{R}^{p_k^{(1)}\times d}$ is the projection matrix and $\mathbf{b}_k^{(1)}\in\mathbb{R}^{p_k^{(1)}}$ is the bias vector to be learned in the first layer of the $k$th network, and $\phi : \mathbb{R}\mapsto\mathbb{R}$ is a nonlinear activation function, \eg, \verb'Relu' or \verb'tanh' function. Then, the output of the first layer of this network is used as the input of the second layer. Therefore, the output of the second layer is $\mathbf{h}_k^{(2)} = \phi(\mathbf{W}_k^{(2)}\mathbf{h}_k^{(1)}+\mathbf{b}_k^{(2)})$, where $\mathbf{W}_k^{(2)}\in\mathbb{R}^{p_k^{(2)}\times p_k^{(1)}}$ is the projection matrix and $\mathbf{b}_k^{(2)}\in\mathbb{R}^{p_k^{(2)}}$ is the bias vector to be learned in the second layer of the $k$th network, respectively. Similarly, the output for the $m_k$th layer is $\mathbf{h}_k^{(m_k)} = \phi(\mathbf{W}_k^{(m_k)}\mathbf{h}_k^{(m_k-1)}+\mathbf{b}_k^{(m_k)})$, and for the output layer is:
\begin{eqnarray}
	\label{eq:1} \mathbf{h}_k^{(M_k)} = \phi\big(\mathbf{W}_k^{(M_k)}\mathbf{h}_k^{(M_k-1)}+\mathbf{b}_k^{(M_k)}\big)
\end{eqnarray}
where $\mathbf{W}_k^{(M_k)}$ is the projection matrix and $\mathbf{b}_k^{(M_k)}$ is the bias vector to be learned for the output layer of the $k$th network, respectively.

For each pair of person samples $\mathbf{x}_i$ and $\mathbf{x}_j$, they can be represented as $f_k(\mathbf{x}_i) = \mathbf{h}_{ik}^{(M_k)}$ and $f_k(\mathbf{x}_j) = \mathbf{h}_{jk}^{(M_k)}$ at the output layer of the $k$th network, and their distance metric of the $k$th network can be measured by computing the squared Euclidean distance between the most top level representations, which is defined as follows:
\begin{eqnarray}
	\label{eq:2} \mathbf{\delta}_k^2(\mathbf{x}_i,\mathbf{x}_j) = \big \|f_k(\mathbf{x}_i) - f_k(\mathbf{x}_j) \big \|_2^2
\end{eqnarray}
The final distance metric of each image pair can be computed as:
\begin{eqnarray}
	\label{eq:3} \mathcal{D}({\mathbf{x}_i,\mathbf{x}_j}) = \sum_{k=0}^{K}w_k(\mathbf{x}_i,\mathbf{x}_j)\mathbf{\delta}_k^2(\mathbf{x}_i,\mathbf{x}_j)
\end{eqnarray}
where $\delta_0$ is the global metric which presents the similarity/dissimilarity common to the whole input space and the metrics $\mathbf{\delta}_k$ where $1<k<K$ exploit the variations over local regions defined by clustering. The $w_k(\mathbf{x}_i,\mathbf{x}_j)$ are nonnegative weights which vary smoothly across the input space, and they are defined as:
\begin{eqnarray}
	\label{eq:4}
	w_k\big(\mathbf{x}_i,\mathbf{x}_j\big)=
	\begin{cases}
	\beta & k=0 \\
	s_k(\mathbf{x}_i)\cdot s_k(\mathbf{x}_j) & \text{otherwise}
	\end{cases}	
\end{eqnarray}
$\beta$ is a positive constant and $s_k(\mathbf{x})$ is the similarity function which presents how close an instance $\mathbf{x}$ is to the $k$th local region.

By adding the weights $w_k(\mathbf{x}_i,\mathbf{x}_j)$, each local region has a strong influence only within a specific smaller space which emphasizes the most discriminative feature. Specifically, $\delta_k$ has a large weight in $\mathcal{D}({\mathbf{x}_i,\mathbf{x}_j})$ if $\mathbf{x}_i$ and $\mathbf{x}_j$ are closely associated with the $k$th local region.

Since the L1-Norm of the vector $(w_1,w_2,\cdots,w_k)$ is scaled to $1$ after regularization, our model is a generalization of global metric and local metrics as the parameter $\beta$ is allowed to balance the influence of the global metric $\delta_0$ and the local metrics $\delta_k$ $(1\leq k \leq K)$ in the matrix $\mathcal{D}({\mathbf{x}_i,\mathbf{x}_j})$. When $K$ is larger, the model will be better at exploiting subtle local variations, while at the same time, be more possible to overfit. Since speed and memory consumption of the method will grow linearly with $K$ while training and testing, we should choose the smallest $K$ if several different values perform comparably. Generally, if $K = 0$ or $\beta \rightarrow \infty$ our model is equivalent to a global nonlinear metric. The impact of $K$ on the performance will be studied in Section~\ref{section:analysis}.

\subsection{Local Distribution}
As shown in Figure~\ref{fig:comparison}, local distributions are obtained through clustering. To include soft $s$, we use the K-means algorithm to get the cluster centers $\mathbf{V}$ and apply the radial basis function\footnote{Some algorithms like the gaussian mixture model or sparse coding can also cluster the data. However, the gaussian mixture model is suitable for low dimensional feature and sparse coding runs inefficiently on large dataset.} as the similarity function in (\ref{eq:4}) which is defined as follows:
\begin{eqnarray}
	\label{eq:5} s_k(\mathbf{x})= \exp\Big({\frac{-\big\|\mathbf{x}-\mathbf{v}_k\big\|^2}{2\sigma^2}}\Big)
\end{eqnarray}
where $\mathbf{V}\in\mathbb{R}^{d\times K}$ is the cluster centers. By doing this, we are capable of representing the local variations in a smooth way and not being restricted to assign an instance to a single local region.

\subsection{Objective Function}
To exploit discriminative information from the final representations with the proposed NLML model, we expect that there is a large margin between positive pairs and negative pairs. Specifically, NLML aims to pursue the nonlinear mapping function $\mathcal{D}$ such that the final distance $\mathcal{D}({\mathbf{x}_i,\mathbf{x}_j})$ between $\mathbf{x}_i$ and $\mathbf{x}_j$ is smaller than a preset parameter $\tau_1$ if $\mathbf{x}_i$ and $\mathbf{x}_j$ are from the same subject and conversely larger than $\tau_2$ if they are from different subjects. The formulation can be represented as follows:
\begin{eqnarray}
	\label{eq:6}\mathcal{D}({\mathbf{x}_i,\mathbf{x}_j}) \leq \tau_1, &y_{ij} = 1 \\
	\label{eq:7}\mathcal{D}({\mathbf{x}_i,\mathbf{x}_j}) \geq \tau_2, &y_{ij} = -1
\end{eqnarray}
where $\tau_1 < \tau_2$ and $y_{ij}$ denotes the similarity or dissimilarity between a person image pair $\mathbf{x}_i$ and $\mathbf{x}_j$.

Presetting $\tau_1$ and $\tau_2$ as $\tau-c$ $(\tau>c)$ and $\tau+c$, we can reduce the parameter and enforce the margin between $\mathcal{D}({\mathbf{x}_i,\mathbf{x}_j})$ by using the following constraint:
\begin{eqnarray}
	\label{eq:8}c-y_{ij}\big(\tau-\mathcal{D}({\mathbf{x}_i,\mathbf{x}_j})\big)<0	
\end{eqnarray}

By applying the constraint in ~(\ref{eq:8}) to each positive pair and negative pair in the training set, we formulate our optimization problem as follows:
\begin{eqnarray}
\label{eq:9} \arg \underset{f}\min~~J &=& J_{1} + J_{2} \nonumber \\
&=& \frac{1}{2}\sum\limits_{i,j} g \Big(c-y_{ij} \big(\tau - \mathcal{D}({\mathbf{x}_i,\mathbf{x}_j}) \big) \Big)   \\
&+& \frac{\lambda}{2} \sum\limits_{k=0}^{K}\sum\limits_{m=1}^{M_k} \Big( \big\|\mathbf{W}_k^{(m_k)} \big \|_F^2 ~+~ \big\|\mathbf{b}_k^{(m_k)} \big \|_2^2 \Big) \nonumber
\end{eqnarray}
where $J_1$ is the logistic loss which force the distances between positive pairs smaller than the distances between negative pairs and $J_2$ regularizes the parameters of the $K+1$ networks, $\lambda$ is the parameter that balance the contribution of different terms and $g(z)$ is the generalized logistic function to approximate the hinge loss function $u = max(u,0)$, and is defined as follows:
\begin{eqnarray}
	\label{eq:10} g(z) = \frac{1}{\gamma}\log\big(1+\exp(\gamma z)\big)
\end{eqnarray}
where $\gamma$ is the sharpness parameter.
\subsection{Optimization}
To solve the optimization problem in (\ref{eq:9}), we employ the batch gradient descent scheme to obtain the parameters $\{\mathbf{W}_k^{(m_k)}, \mathbf{b}_k^{(m_k)}\}$, $m_k = 1, 2, \cdots, M_k$ and $k = 0,1,2,\cdots,K$. The gradients of the objective function $J$ with respect to $\mathbf{W}_k^{(m_k)}$ and $\mathbf{b}_k^{(m_k)}$ can be computed as follows:

\begin{eqnarray}
\label{eq:11}   \frac{\partial{J}}{\partial{\mathbf{W}_k^{(m_k)}}} &=& \sum\limits_{i,j} \Big( \mathbf{\Psi}_{k,ij}^{(m_k)} {\mathbf{h}_{k,i}^{(m_k-1)}}^T + \mathbf{\Psi}_{k,ji}^{(m_k)} {\mathbf{h}_{k,j}^{(m_k-1)}}^T \Big) \nonumber  \\
 &+& \lambda~ \mathbf{W}_k^{(m_k)}  \\
\label{eq:12}  \frac{\partial{J}}{\partial{\mathbf{b}_k^{(m_k)}}} &=& \sum\limits_{i,j} \Big( \mathbf{\Psi}_{k,ij}^{(m_k)} + \mathbf{\Psi}_{k,ji}^{(m_k)} \Big) +  \lambda ~\mathbf{b}_k^{(m_k)}
\end{eqnarray}
where $\mathbf{\Psi}_{k,ij}$ and $\mathbf{\Psi}_{k,ji}$ are two updating functions. For the output layer $(m_k = M_k)$ of each network, they are computed as follows:
\begin{eqnarray}
\nonumber
\label{eq:13} \mathbf{\Psi}_{k,ij}^{(M_k)} &=& w_k(\mathbf{x}_i,\mathbf{x}_j)g'(e) \big(\mathbf{h}_{k,i}^{(M_k)}-\mathbf{h}_{k,j}^{(M_k)} \big)  \odot \phi' \big(\mathbf{z}_{k,i}^{(M_k)} \big) \\  \nonumber
\label{eq:14} \mathbf{\Psi}_{k,ji}^{(M_k)} &=& w_k(\mathbf{x}_i,\mathbf{x}_j)g'(e) \big(\mathbf{h}_{k,j}^{(M_k)}-\mathbf{h}_{k,i}^{(M_k)} \big)  \odot \phi' \big(\mathbf{z}_{k,j}^{(M_k)} \big) 
\end{eqnarray}
where
\begin{eqnarray}
\label{eq:15} e &\triangleq&  c-y_{ij} \big(\tau - \mathcal{D}({\mathbf{x}_i,\mathbf{x}_j}) \big) \\
\label{eq:16} \mathbf{z}_{k,i}^{(m_k)} &\triangleq& \mathbf{W}_k^{(m_k)} {\mathbf{h}_{k,i}^{(m_k-1)}} + \mathbf{b}_k^{(m_k)}\\
\label{eq:17} \mathbf{z}_{k,j}^{(m_k)} &\triangleq& \mathbf{W}_k^{(m_k)} {\mathbf{h}_{k,j}^{(m_k-1)}} + \mathbf{b}_k^{(m_k)}
\end{eqnarray}

For all other layers $(1 \leq m_k \leq M_k)$ of the network, $\mathbf{\Psi}_{k,ij}$ and $\mathbf{\Psi}_{k,ji}$ are computed as follows:
\begin{eqnarray}
\label{eq:18} \mathbf{\Psi}_{k,ij}^{(m_k)} &=& \big({\mathbf{W}_k^{(m_k+1)}}^T \mathbf{\Psi}_{k,ij}^{(m_k+1)} \big) \odot \phi' \big(\mathbf{z}_{k,i}^{(m_k)} \big) \\
\label{eq:19} \mathbf{\Psi}_{k,ji}^{(m_k)} &=& \big({\mathbf{W}_k^{(m_k+1)}}^T \mathbf{\Psi}_{k,ji}^{(m_k+1)} \big) \odot \phi' \big(\mathbf{z}_{k,j}^{(m_k)} \big)
\end{eqnarray}
where the operation $\odot$ denotes the element-wise multiplication.

Then, we can use the following gradient descent algorithm to update the $\mathbf{W}_k^{(m_k)}$ and $\mathbf{b}_k^{(m_k)}$ until convergence:

\begin{eqnarray}
\label{eq:20}   \mathbf{W}_k^{(m_k)} &=& \mathbf{W}_k^{(m_k)} - \mu \frac{\partial{J}}{\partial{\mathbf{W}_k^{(m_k)}}}\\
\label{eq:21}   \mathbf{b}_k^{(m_k)} &=& \mathbf{b}_k^{(m_k)} - \mu \frac{\partial{J}}{\partial{\mathbf{b}_k^{(m_k)}}}
\end{eqnarray}
where $\mu$ is the learning rate.

In practice, we apply the greedy layer-wise algorithm to pre-train the networks then fine-tuning the parameters with smaller learning rate $\mu$. An EM-like iterative optimization algorithm is utilized to alternatively optimize the $\mathbf{V}$ in clustering and the networks. Both steps decrease the objective function and achieve the convergence. \textbf{Algorithm~\ref{alg:1}} summarizes the detailed procedure of the proposed NLML method.
\begin{algorithm}[tb]
\DontPrintSemicolon
\KwIn{Training set: $\mathbf{X}$, number of local regions $K$, network layer number $M_k+1$, learning rate $\mu$, iterative number $T$, parameter $\lambda$, threshold $\tau$, margin $c$ and convergence error $\varepsilon$.}
\KwOut{Parameter $\mathbf{W}_k^{m_k}$ and $\mathbf{b}_k^{m_k}$, $1\leq m_k \leq M_k$, $0\leq k\leq K$}
\SetAlgoLined
Initialize $\mathbf{W}_k^{(m_k)}$ and $\mathbf{b}_k^{(m_k)}$ with proper value. \;
Cluster data and get $K$ local regions by K-means.\;
Obtain $s_k$ for each image according to ~(\ref{eq:5}).\;
\For{$t = 1, 2, \cdots, T$}{
Do forward propagation to all the training samples \;
\For{$m_k = 1, 2, \cdots, M_k$}{
Get $\mathbf{h}_{k,i}^{(m_k)}$ and $\mathbf{h}_{k,j}^{(m_k)}$ by forward propagation.}
\For{$m_k = M_k, M_k-1, \cdots, 1$}{
Obtain gradient by back propagation according to ~(\ref{eq:11}) and (\ref{eq:12}).}
\For{$m_k = 1, 2, \cdots, M_k$}{
Update $\mathbf{W}_k^{(m_k)}$ and $\mathbf{b}_k^{(m_k)}$ according to ~(\ref{eq:20}) and (\ref{eq:21}).}
Calculate $J_t$ using ~(\ref{eq:9}). \;
If $t > 1$ and $|J_{t}-J_{t-1}|<\varepsilon$, go to \textbf{Return}.\;}
\textbf{Return:} $\mathbf{W}_k^{m_k}$ and $\mathbf{b}_k^{m_k}$, $1\leq m_k \leq M_k$, $0\leq k\leq K$.
\caption{NLML} \label{alg:1}
\end{algorithm}

\section{Experiments}

We evaluated our approach by conducting person re-identification experiments on three widely-used datasets: VIPeR~\cite{gray2007evaluating}, GRID~\cite{loy2009multi}, and CUHK 01~\cite{li2012human}. We described the details of the experiments and results in the following.

\subsection{Experiments on VIPeR}

The VIPeR dataset contains 632 person im.pdfage.pdf pairs captured from two different views. All images are scaled to $128 \times 48$ pixels. The dataset is especially challenging for two main reasons: 1) viewpoint changes for most image pairs are near or over 90 degrees and this makes it difficult to associate the person from two views; 2) image resolution is much lower compared with images of other datasets.

\begin{figure*}
\centering
\begin{minipage}[b]{0.4\linewidth}
    \includegraphics[width=0.95\textwidth]{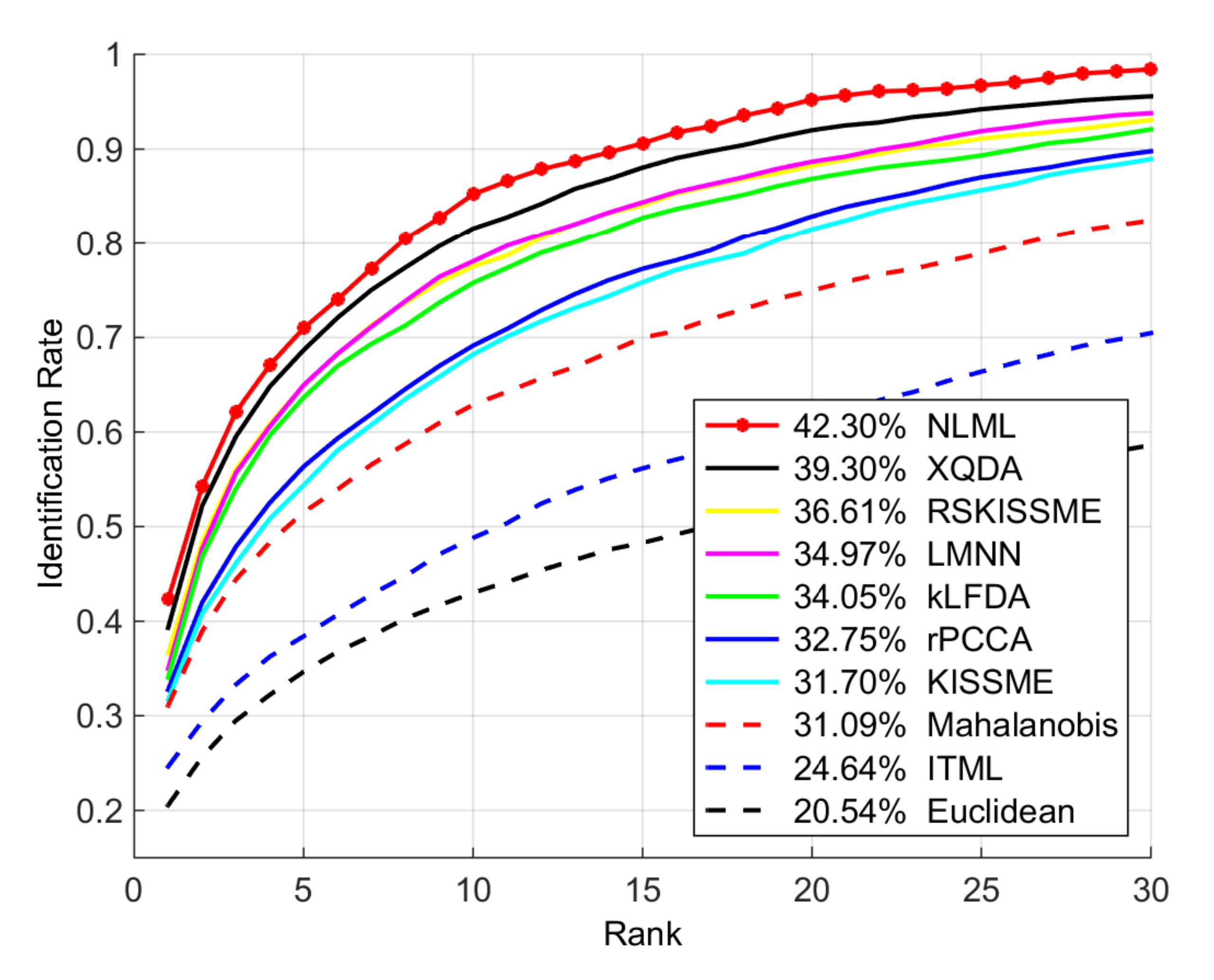}
    \caption*{(a) VIPeR (p=316)}
\end{minipage}
\begin{minipage}[b]{0.4\linewidth}
    \includegraphics[width=0.95\textwidth]{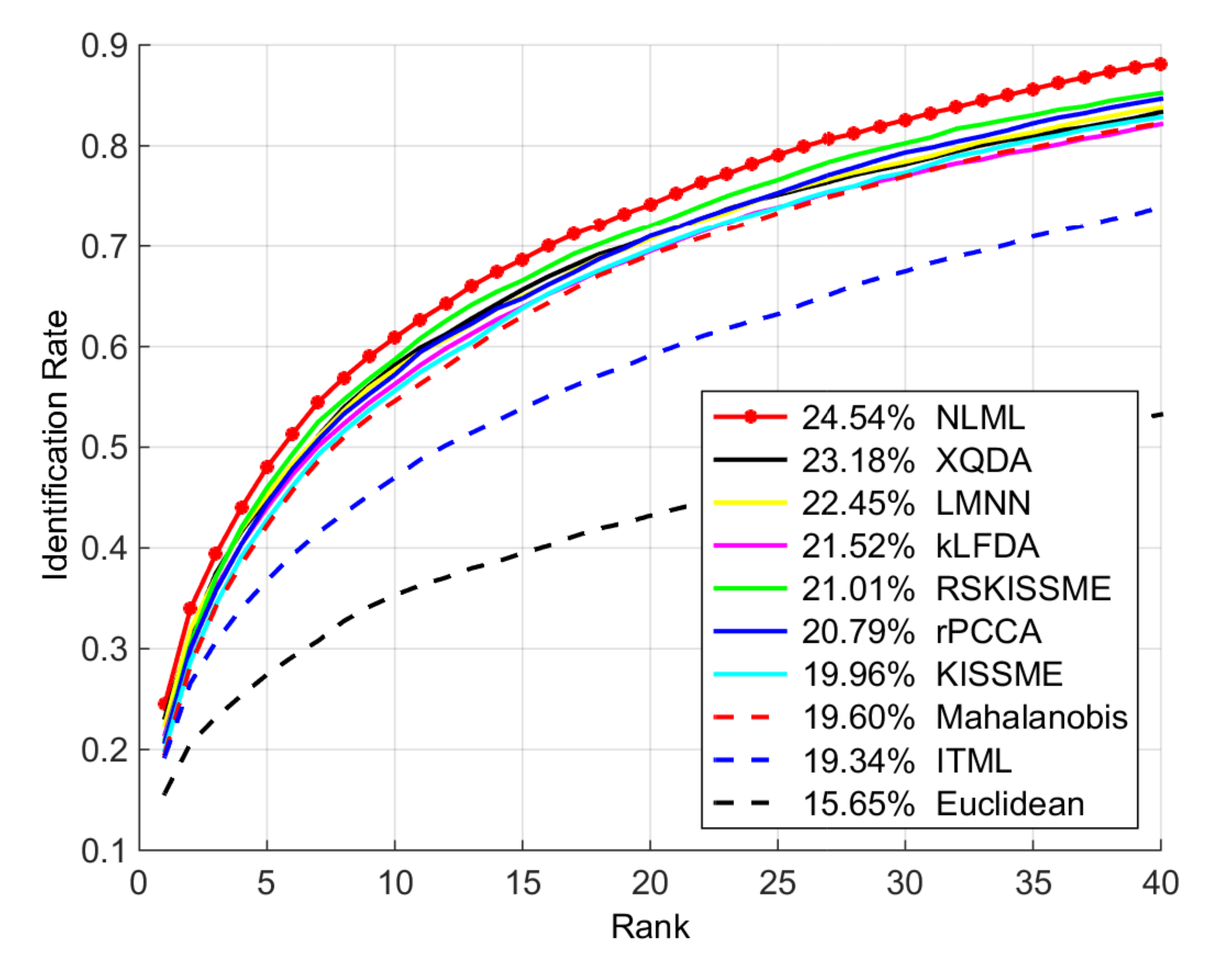}
    \caption*{(b) VIPeR (p=100)}
\end{minipage}
\caption{Comparison of different metric learning methods on the VIPeR dataset~\cite{gray2007evaluating}. (a) The size of gallery set is 316. (b) The size of gallery set is 100. }
\label{fig:viper_result}
\end{figure*}
\vspace{1pt}
\begin{figure*}
\centering
\begin{minipage}[b]{0.4\linewidth}
    \includegraphics[width=0.95\textwidth]{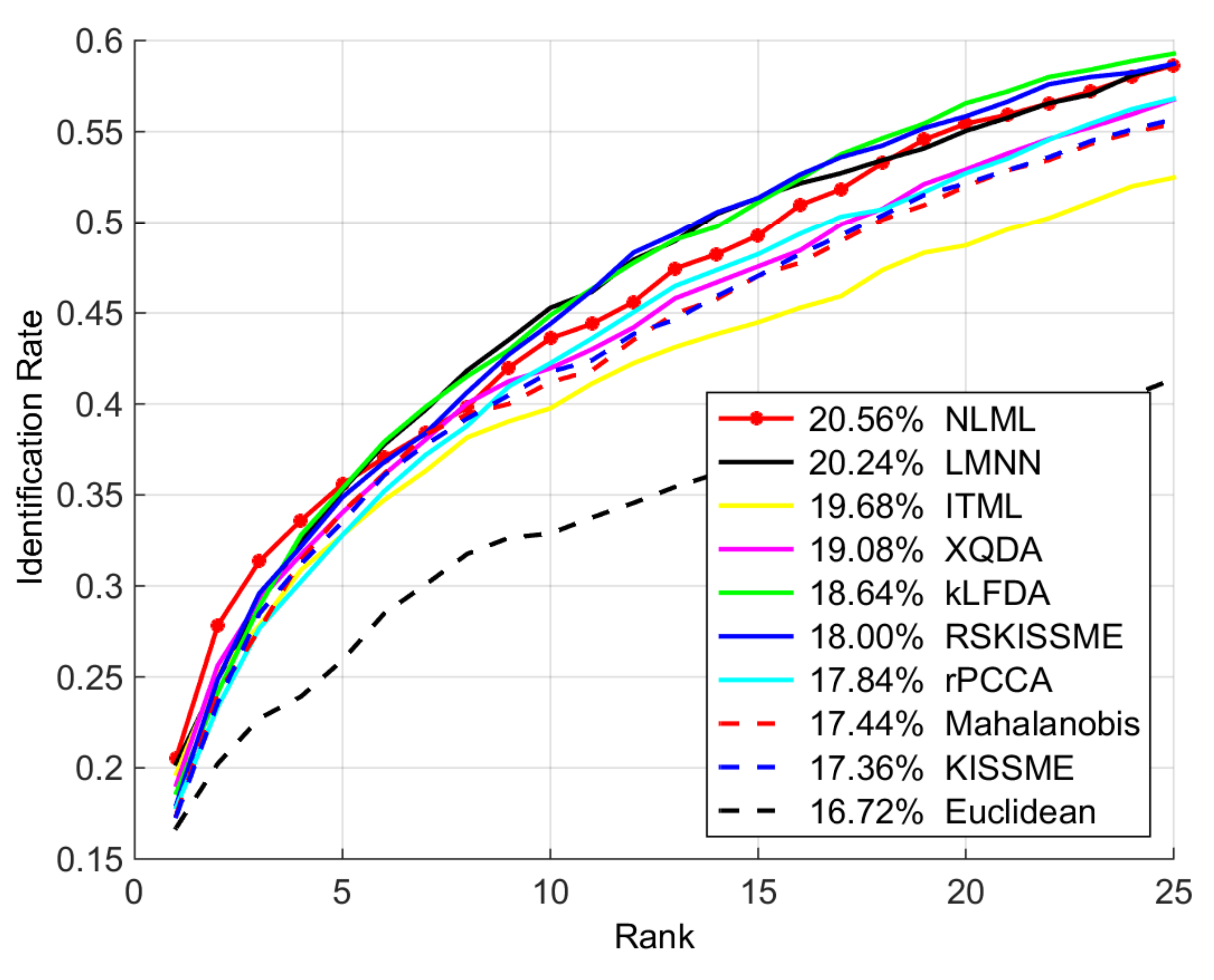}
    \caption*{(a) GRID (LOMO)}
\end{minipage}
\begin{minipage}[b]{0.4\linewidth} 
    \includegraphics[width=0.95\textwidth]{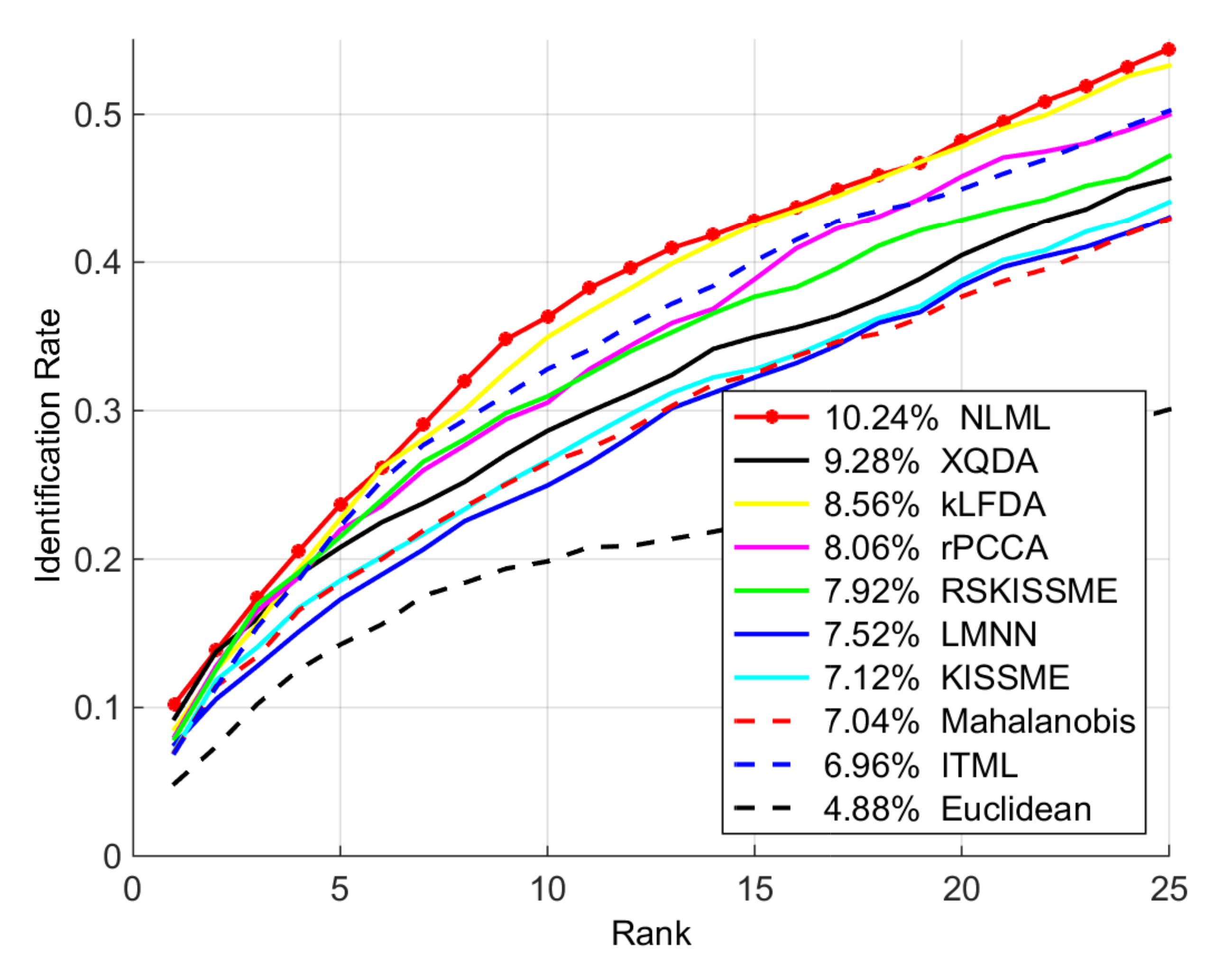}
    \caption*{(b) GRID (ELF6)}
\end{minipage}
\caption{Comparison of different metric learning methods on the GRID dataset~\cite{loy2009multi} (p=900). (a) CMC curves with LOMO feature. (b) CMC curves with ELF6 feature.}
\label{fig:grid_result}
\end{figure*}

\textbf{Evaluation protocols}: Two widely adopted experimental protocols were used for the evaluation: randomly dividing the 632 pairs of image pairs into half (316 image pairs) for training and the other half for testing; randomly selecting 100 image pairs as the training set and the remaining 532 image pairs as the testing set. We adopted the single-shot evaluation and partition the testing set into probe set and the gallery set. The result is evaluated by cumulative matching characteristic (CMC) curve~\cite{gray2007evaluating}, an estimate of finding the correct match in the top $n$ match. Final performance is averaged over ten random repeats of the process.

\textbf{Feature representation}: We used the LOMO feature for pedestrian image representation~\cite{liao2015person}. It extracts Scale Invariant Local Ternary Pattern (SILTP)~\cite{liao2010modeling} and HSV features from the image to constitute high-level descriptor. This descriptor is better at solving the person re-identification problem since it applies the Retinex transformation and a scale invariant texture operator, analyzes the horizontal occurrence of local features and maximizes the occurrence to handle viewpoint changes. In the training stage, we horizontal reflected each training image to increase the number of positive pairs.

\textbf{Parameter settings}: For our NLML method, we clustered 4 local regions in VIPeR by setting $K = 4$ after testing the performance with varying $K$ (see Section~\ref{section:analysis}). We needed to train 1 global network and 4 local networks and designed each network with 3 layers where $M_k = 3, k = 0,1,2,3,4$. The dimensions for these layers were empirically set as 500, 400 and 300, respectively. The global weight $\beta$, learning rate $\mu$, parameter $\lambda$, threshold $\tau$, margin $c$ and convergence error $\epsilon$ were empirically set as 1, 0.004, 0.01, 2, 1, 0.1, respectively. The parameters $\mathbf{W}_k^{m_k}$ of our NLML model were initialized as $\mathbf{I}^{(p_{m_k-1})\times (p_{m_k})}$ ($p_{m_k}$ is the feature dimension of the $m_k$th layer), which is a matrix with ones on the diagonal and zeros elsewhere. The bias vector $\mathbf{b}_k^{m_k}$ was initialized as zero vectors. For the activation function, we used the non-saturating sigmoid function in our experiments. Except the number of local cluster $K$ and the global weight $\beta$, this parameter setting was also applied to the other datasets.
	
\textbf{Comparison with existing metric learning algorithms}: We compared the proposed NLML algorithm with several existing metric learning algorithms, including the Euclidean distance and the Mahalanobis distance which are two baseline approaches applied in person re-identification problem, and some existing state-of-the-art algorithms such as LMNN~\cite{weinberger2005distance}, ITML~\cite{davis2007information}, KISSME~\cite{koestinger2012large},  kLFDA~\cite{xiong2014person} and XQDA~\cite{liao2015person}. In order to remove the redundancy of the high-dimensional feature space and achieve fair comparison, we first applied PCA to reduce the feature dimensionality from 26960 to 500. Figure~\ref{fig:viper_result} shows that our NLML method outperforms most existing metric learning methods. It achieved 42.30\% and 24.30\% rank-1 accuracy for p = 316 and p = 100 perspectively.

\textbf{Comparison with the state-of-the-art methods}: We also compared the performance of our NLML method with the state-of-the-art results reported on the VIPeR dataset which applied same protocols such as XQDA~\cite{liao2015person}, PolyMap~\cite{chen2015similarity}, kLFDA~\cite{xiong2014person}, kBiCov~\cite{ma2014covariance} and SalMatch~\cite{zhao2013person}. We observe from Table~\ref{tab:viper_316} and ~\ref{tab:viper_100} that our method obtains better performance than current state-of-the-art. Specifically, our NLML method outperforms the second-best, LOMO+XQDA, by 2.20\% at rank 1 when p is set as 316 and improves the second-best method PolyMap with 3.14\% when p is set as 100.

\begin{table}[tb]\small
\centering
\caption{Matching rates (\%) of different state-of-the-art person re-identification methods on the VIPeR dataset (p=316).}
\label{tab:viper_316}
\begin{tabular}{|l|c|c|c|c|} \hline
\textbf{Method}                  & \textbf{r = 1} & \textbf{r = 5} & \textbf{r = 10} & \textbf{r = 20}    \\ \hline
NLML (ours)    & \textbf{42.30} & \textbf{70.99} & \textbf{85.23} & \textbf{94.25}     \\
LOMO+XQDA~\cite{liao2015person} & 40.00 & 68.13 & 80.51 & 91.08 \\
PolyMap~\cite{chen2015similarity} & 36.80 & 70.40 & 83.70 & 91.70 \\
kLFDA~\cite{xiong2014person} & 32.30 & 65.80 & 79.70 & 90.90 \\
kBiCov~\cite{ma2014covariance} & 31.11 & 58.33 & 70.71 & 82.44 \\
SalMatch~\cite{zhao2013person} & 30.16 & 52.31 & 75.31 & 86.71 \\
LADF~\cite{li2013learning} & 29.88 & 61.04 & 75.98 & 88.10 \\
MidFilter~\cite{zhao2014learning} & 29.11 & 52.34 & 65.95 & 78.80 \\
McMCML~\cite{ma2014person} & 28.83 & 59.34 & 75.82 & 88.51 \\
LFDA~\cite{pedagadi2013local} & 24.18  & 52.00 & 67.12 & 82.00 \\
eLDFV~\cite{ma2012local} & 22.34 & 47.00 & 67.04 & 71.00 \\
KISSME~\cite{koestinger2012large} & 19.60 & 35.00 & 62.20 & 77.00 \\
PCCA~\cite{mignon2012pcca} & 19.27 & 48.89 & 64.91 & 80.28 \\
PRDC~\cite{zheng2013reidentification} & 15.66 & 38.42 & 53.86 & 70.09 \\
ELF~\cite{gray2008viewpoint} & 12.00 & 31.20 & 41.00 & 58.00 \\
\hline
\end{tabular}
\end{table}

\begin{table}[tb]\small
\centering
\caption{Matching rates (\%) of different state-of-the-art person re-identification methods on the VIPeR dataset (p=100).}
\label{tab:viper_100}
\begin{tabular}{|l|c|c|c|c|} \hline
\textbf{Method}                  & \textbf{r = 1} & \textbf{r = 5} & \textbf{r = 10} & \textbf{r = 20}    \\ \hline
NLML (ours)    & \textbf{20.56} & \textbf{48.02} &\textbf{60.88} & \textbf{74.15} \\
PolyMap~\cite{chen2015similarity} & 17.40 & 41.60 & 55.30 & 70.80 \\
LADF~\cite{li2013learning}  & 12.90 & 30.30 & 42.70 & 58.00 \\
MtMCML~\cite{ma2014person} & 12.33 & 31.64 & 45.13 & 61.11 \\
RPML~\cite{hirzer2012relaxed} & 10.90 & 26.70 & 37.70 & 51.60 \\
PCCA ~\cite{mignon2012pcca} & 9.27 & 24.89 & 37.43 & 52.89 \\
PRDC~\cite{zheng2013reidentification} & 9.12 & 24.19 & 34.40 & 48.55 \\
\hline
\end{tabular}
\end{table}

\begin{table}[tb]\small
\centering
\caption{Matching rates (\%) of different state-of-the-art person re-identification methods on the GRID dataset (p=900).}
\label{tab:grid_comparison}
\begin{tabular}{|l|c|c|c|c|} \hline
\textbf{Method}                  & \textbf{r = 1} & \textbf{r = 5} & \textbf{r = 10} & \textbf{r = 20}    \\ \hline
NLML (ours)    & \textbf{24.54} & \textbf{35.86} & 43.53 & 55.25     \\
LOMO+XQDA~\cite{liao2015person} & 16.56 & 33.84 & 41.84 & 52.40 \\
PolyMap~\cite{chen2015similarity} & 16.30 & 35.80 & \textbf{46.00} & 57.60 \\
MtMCML~\cite{ma2014person} & 14.08 & 34.64 & 45.84 & \textbf{59.84}\\
MRank-RankSVM~\cite{loy2013person} & 12.24 & 27.84 & 36.32 & 46.56 \\
MRank-PRDC ~\cite{loy2013person} & 11.12 & 26.08 & 35.76 & 46.56 \\
LCRML~\cite{chen2014relevance} & 10.68 & 25.76 & 35.04 & 46.48 \\
XQDA~\cite{liao2015person} & 10.48 & 28.08 & 38.64 & 52.56 \\
RankSVM~\cite{prosser2010person} & 10.24 & 24.56 & 33.28 & 43.68 \\
PRDC~\cite{zheng2013reidentification} & 9.68 & 22.00 & 32.96 & 44.32 \\
L1-Norm\cite{loy2013person} & 4.40 & 11.68 & 16.24 & 24.80 \\
\hline
\end{tabular}
\end{table}

\begin{figure}
\centering
\includegraphics[width=0.35\textwidth]{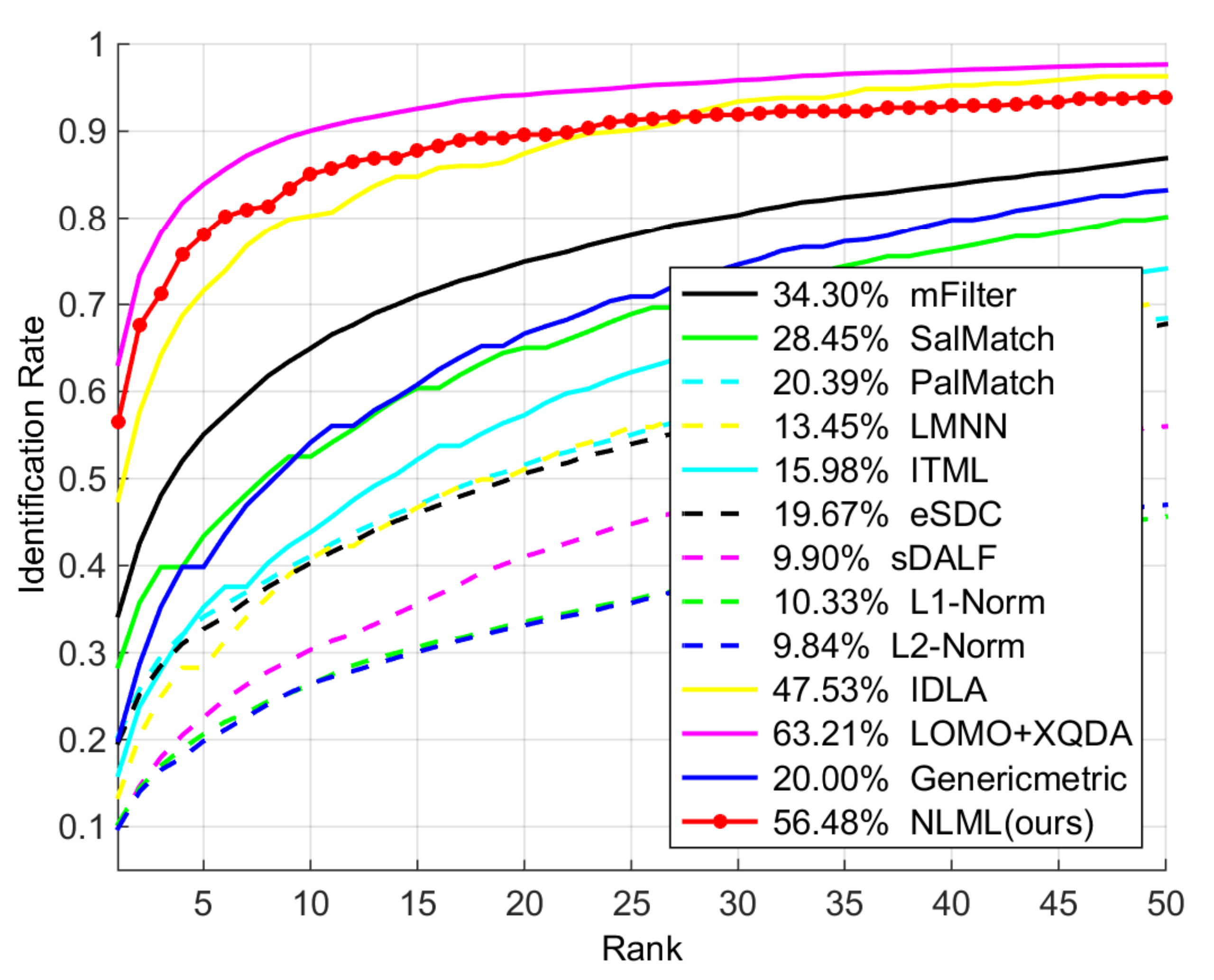}
\caption{Comparison of multi-shot CMC curves and rank-1 identification rates on the CUHK 01 dataset.}
\label{fig:cuhk_result}
\end{figure}

\begin{figure}
\centering
\begin{minipage}[b]{0.48\linewidth}
    \includegraphics[width=\textwidth]{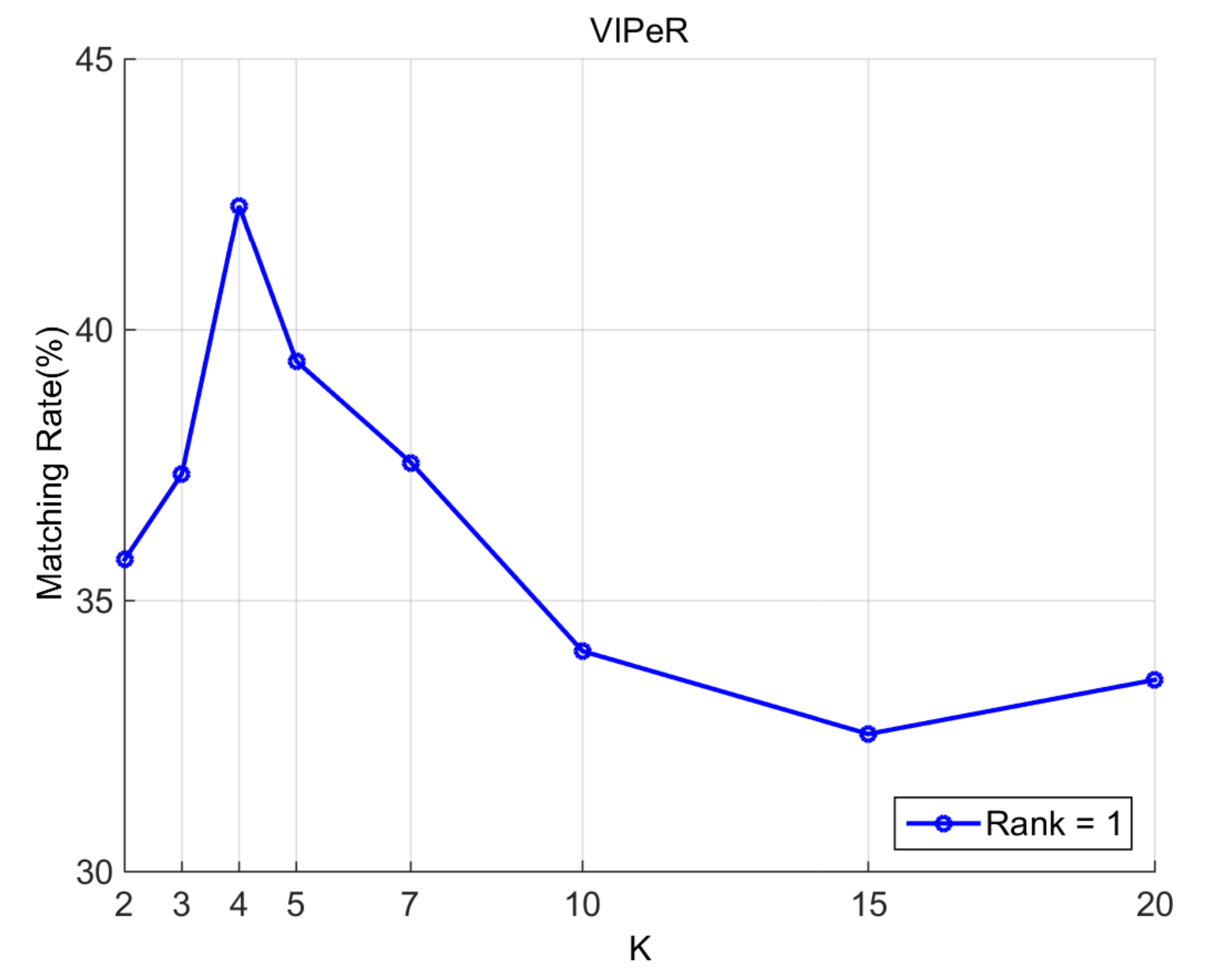}
    \caption*{(a) VIPeR}
\end{minipage}
\begin{minipage}[b]{0.48\linewidth}
    \includegraphics[width=\textwidth]{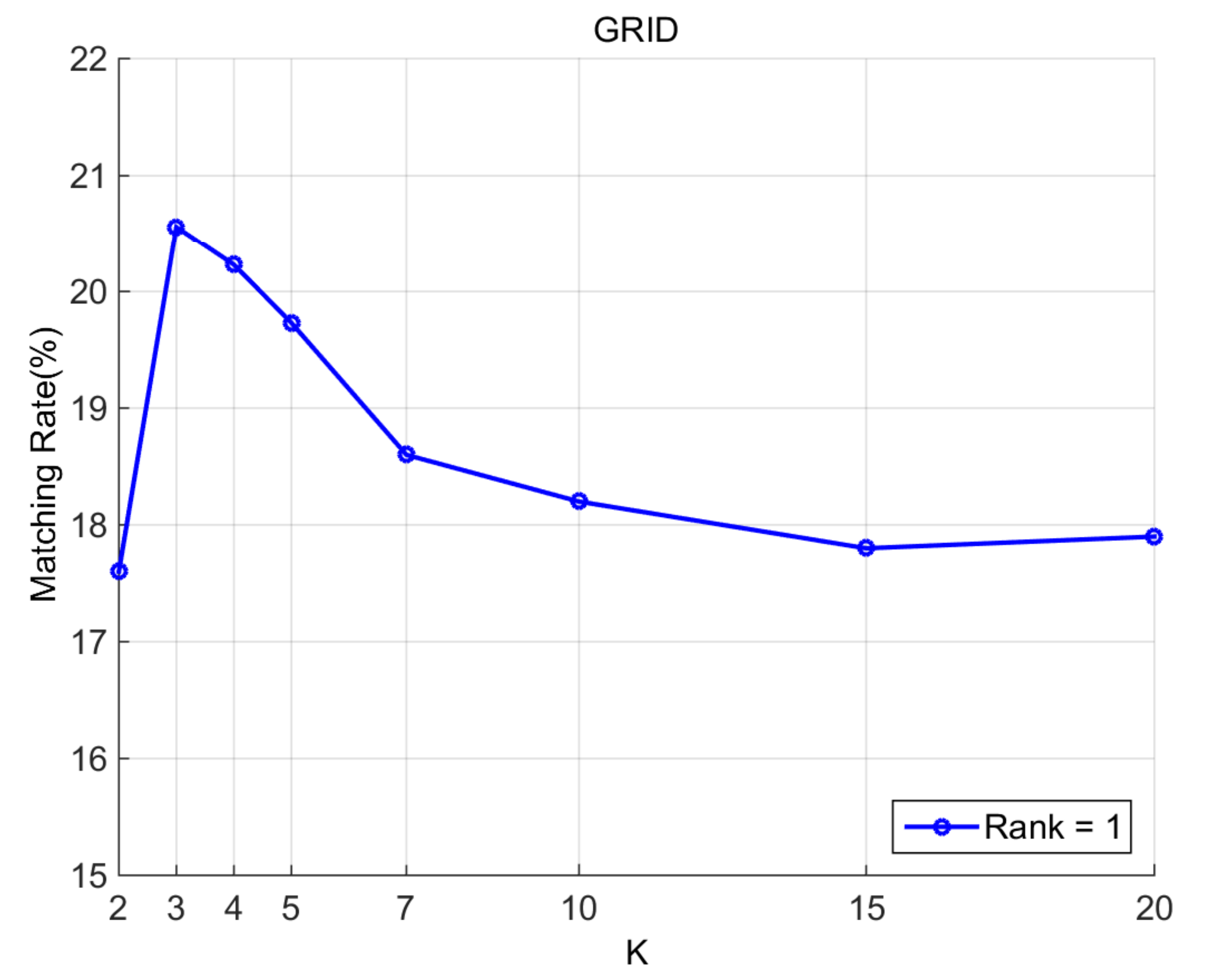}
    \caption*{(b) GRID}
\end{minipage}
\caption{Rank-1 matching rate at varying $K$ on the VIPeR and GRID dataset.}
\label{fig:compare_k}
\end{figure}

\begin{figure}
\centering
\begin{minipage}[b]{0.48\linewidth}
    \includegraphics[width=\textwidth]{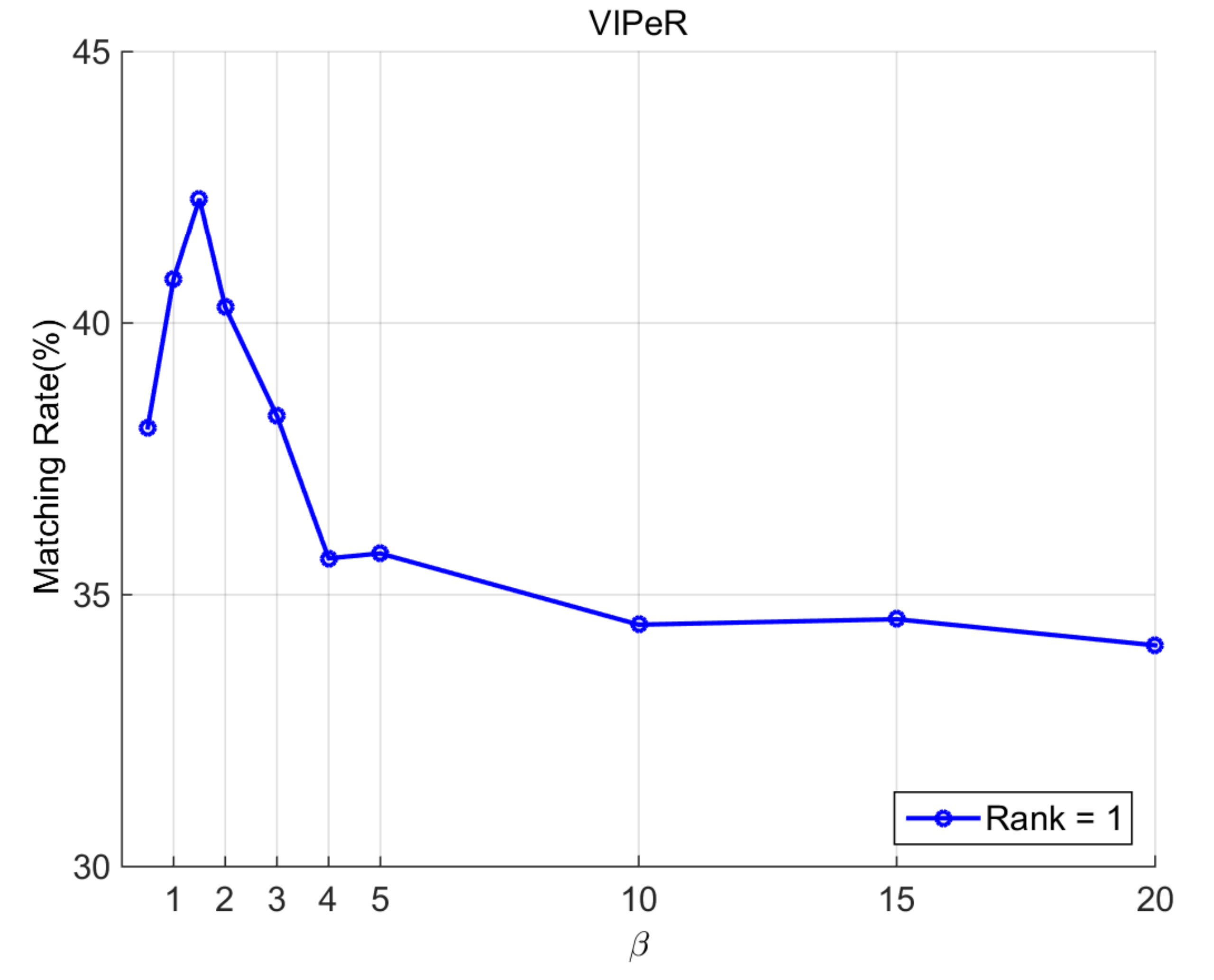}
    \caption*{(a) VIPeR}
\end{minipage}
\begin{minipage}[b]{0.48\linewidth}
    \includegraphics[width=\textwidth]{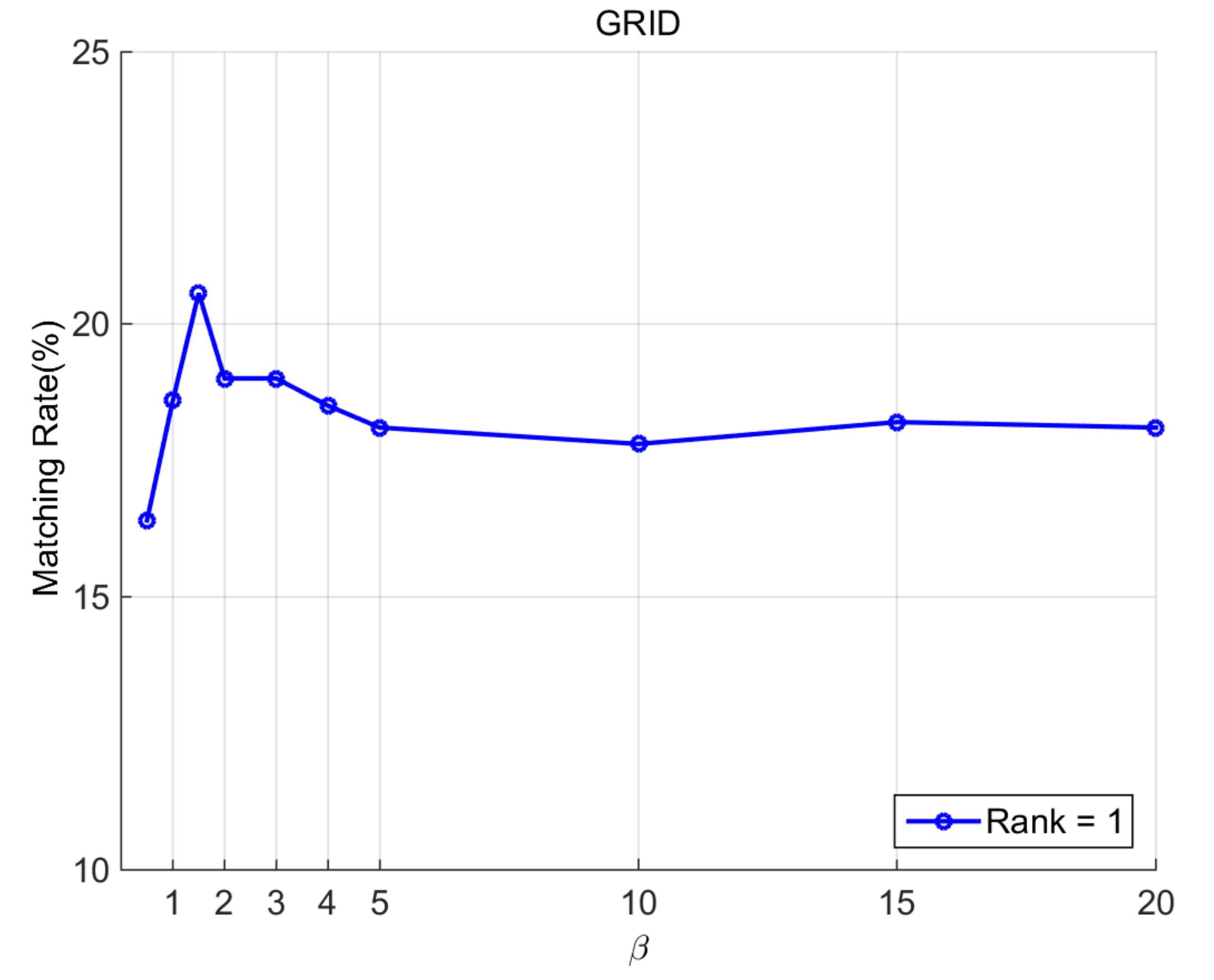}
    \caption*{(b) GRID}
\end{minipage}
\caption{Rank-1 matching rate at varying $\beta$ on the VIPeR and GRID dataset.}
\label{fig:compare_beta}
\end{figure}

\begin{figure*}

\centering
\begin{minipage}[b]{0.30\linewidth}
    \includegraphics[width=0.92\textwidth]{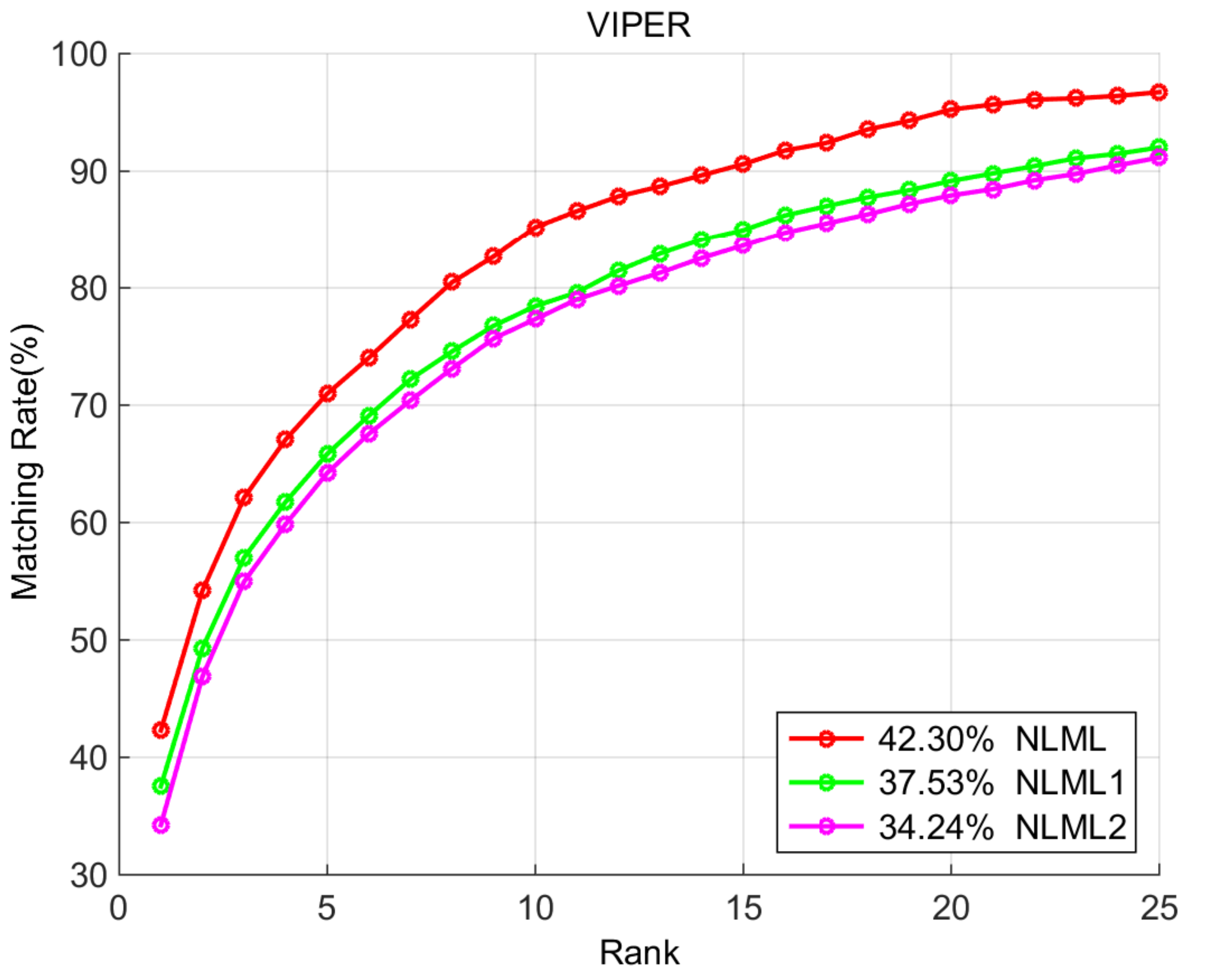}
    \caption*{(a) VIPeR (p = 316)}
\end{minipage}
\begin{minipage}[b]{0.30\linewidth}
    \includegraphics[width=0.92\textwidth]{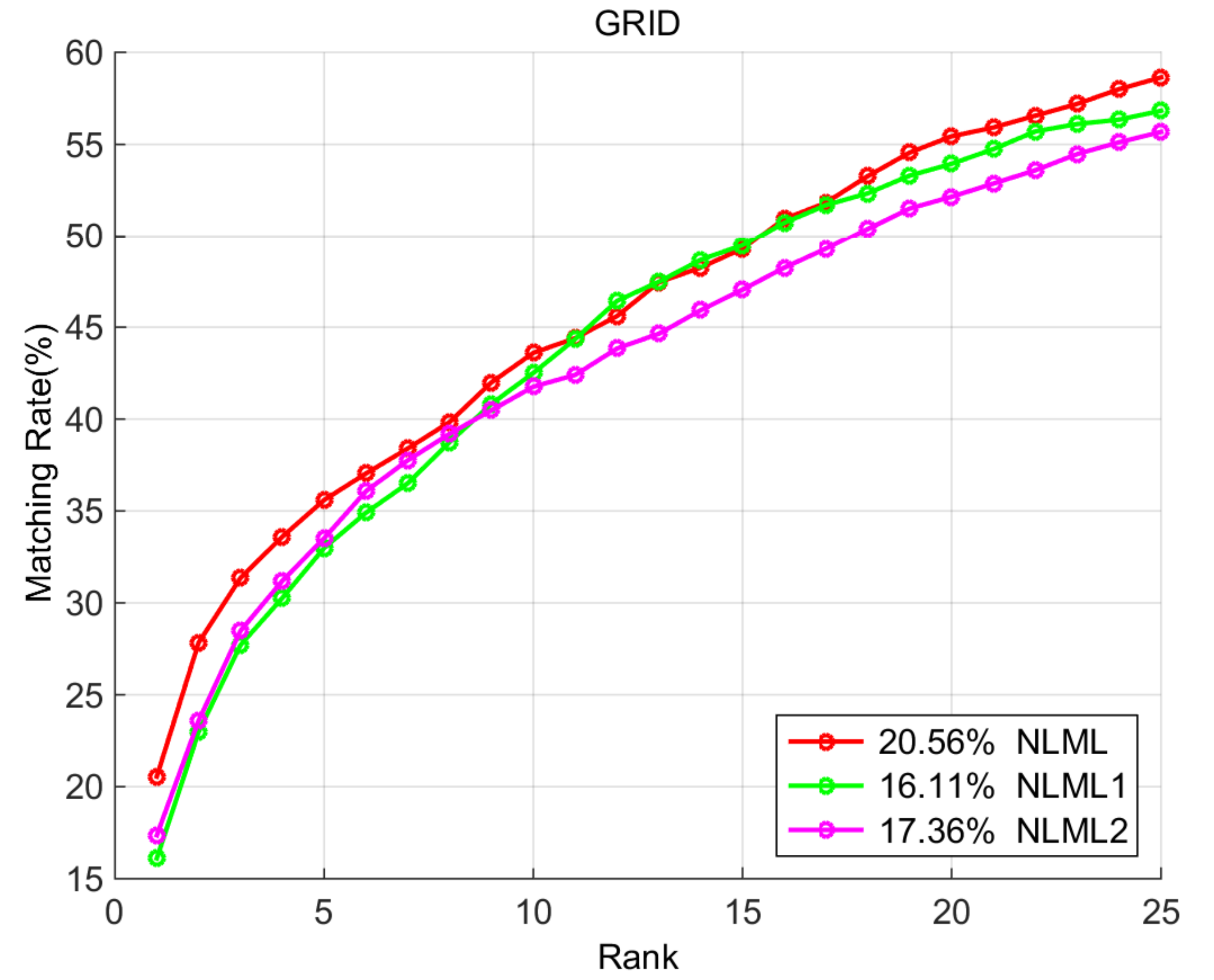}
    \caption*{(b) GRID (p = 900)}
\end{minipage}
\begin{minipage}[b]{0.30\linewidth}
    \includegraphics[width=0.92\textwidth]{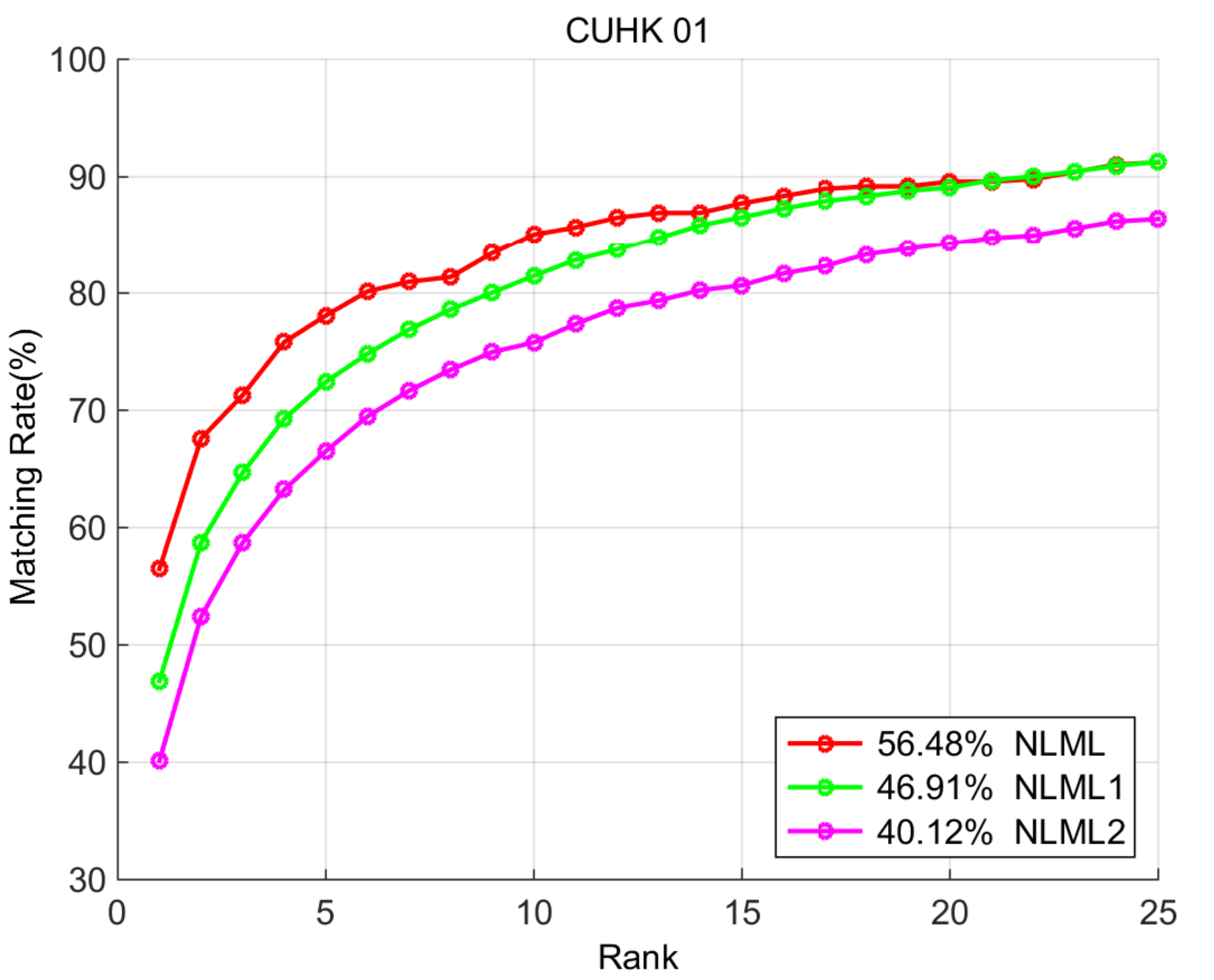}
    \caption*{(c) CUHK 01 (p = 486)}
\end{minipage}
\caption{Matching rates of different variations of our model on the VIPeR, GRID and CUHK 01 dataset. NLML1 represents the model without local metric learning; NLML2 represents the model without nonlinear metric learning.}
\label{fig:local_contribution}
\end{figure*}

\subsection{Experiments on QMUL GRID}
The QMUL underGround Re-IDentnification (GRID) dataset consists of 250 pedestrian image pairs and 775 extra individual images. This is also a difficult person re-identification dataset due to the large viewpoint, pose variations and the low resolution caused by the setting of the camera network which consists of eight disjoint camera views in a busy underground station. An experimental setting of 10 random trials is provided for the GRID dataset. In each trial, 125 image pairs are utilized for training, the remaining 125 image pairs are used for testing, and the extra 775 images constitute part of the gallery set during training.

We first employed the LOMO feature to evaluate our NLML method. We shared the parameter setting of the experiments on VIPeR except that we set the cluster number $K$ as $3$ and the global weight as $1.5$ due to the parameter analysis (see Section~\ref{section:analysis}). Figure~\ref{fig:grid_result} (a) shows our method obtains highest rank-1 matching rate 20.56\% and is comparable with other metric learning methods across all the ranks. The GRID dataset also provides the ELF6~\cite{liu2012person} feature with 2784 dimensions for testing metric learning algorithms and we reduced the feature dimensionality to 500. It can be observed from Figure~\ref{fig:grid_result} (b) that our NLML method obtains better performance than other methods. Table~\ref{tab:grid_comparison} summarizes the state-of-the-art algorithms that reported on the GRID dataset such as XQDA~\cite{liao2015person}, PolyMap~\cite{chen2015similarity}, MtMCML~\cite{ma2014person} and LCRML~\cite{chen2014relevance}. Our NLML achieves the best rank-1 accuracy with LOMO feature.

\subsection{Experiments on CUHK 01}
The CUHK 01 dataset contains 971 persons captured from two camera views in a campus environment. Generally, camera view A captured the frontal or back view of the individuals and camera view B captured the profile views. The images in the dataset are of higher resolution compared with other datasets. We set the number of individuals in the training split to 485 and test split to 486. We extracted the LOMO feature and reduced the feature dimensionality to 500. The cluster number $K$ and global weight $\beta$ was empirically set as $7$ and $1.5$, respectively. The multi-shot evaluation procedure was repeated 10 times and the average of CMC curves across 10 partitions is reported. The results are shown in Figure~\ref{fig:cuhk_result}. We found that our method achieves comparable result with other state-of-the-art algorithms.

\subsection{Analysis}
\label{section:analysis}
\textbf{Parameter analysis}: The number of local metrics $K$ and the global weight $\beta$ in local distribution play an important role on the performance of the NLML method. We analyzed the results with varying $K$ and $\beta$ on the VIPeR and GRID datasets with LOMO feature. Figure~\ref{fig:compare_k} shows the rank-1 matching rate at varying $K$ when $\beta$ was set to $1$ and $2$, respectively. We find that the optimal performance of our method was obtained when $K$ is set to $4$ and $3$, respectively. We also tested the varying $\beta$ by setting the $K$ as $4$ and $3$ for each dataset. Figure~\ref{fig:compare_beta} shows that when $\beta$ is low, our model performs weakly because it gives similar weights to the global weight $\beta$ and local weights $w_k$ after regularization, which results in the problem of overfitting.

\textbf{Contribution of different components in NLML}: We investigated the contribution of local metric learning and nonlinear metric learning. We defined two alternative baselines to study the significance of two metric learning terms: (1) NLML1: learning with only nonlinear metric learning; (2) NLML2: learning with only local metric learning. For NLML1, $K$ is set as $0$ and there is only one global network, we adapted similar optimization method to get the network parameters. For NLML2, the deep neural networks are replaced with Mahalanobis distance metrics. Figure~\ref{fig:local_contribution} shows the comparison on the VIPeR, GRID and CUHK 01 dataset. As shown, NLML outperforms both NLML1 and NLML2 in rank-1 accuracy on three datasets, further indicates that both local metric learning and nonlinear metric learning contribute to the promising performance of our model.

\section{Conclusion}

We have proposed a nonlinear local metric learning (NLML) method for person re-identification in this paper. Experimental results on three widely used challenging datasets demonstrate the effectiveness of the proposed method. An interesting direction for future research is how to employ our NLML to other computer vision tasks such as face verification and image classification.

{
\small
\bibliographystyle{ieee}
\bibliography{egbib}

\begin{thebibliography}{10}\itemsep=-1pt

\bibitem{ahmed2015improved}
E.~Ahmed, M.~Jones, and T.~K. Marks.
\newblock An improved deep learning architecture for person re-identification.
\newblock pages 3908--3916, 2015.

\bibitem{bazzani2012multiple}
L.~Bazzani, M.~Cristani, A.~Perina, and V.~Murino.
\newblock Multiple-shot person re-identification by chromatic and epitomic
  analyses.
\newblock {\em PRL}, pages 898--903, 2012.

\bibitem{bohne2014large}
J.~Bohn{\'e}, Y.~Ying, S.~Gentric, and M.~Pontil.
\newblock Large margin local metric learning.
\newblock In {\em ECCV}, pages 679--694. 2014.

\bibitem{chen2015similarity}
D.~Chen, Z.~Yuan, G.~Hua, N.~Zheng, and J.~Wang.
\newblock Similarity learning on an explicit polynomial kernel feature map for
  person re-identification.
\newblock In {\em CVPR}, pages 1565--1573, 2015.

\bibitem{chen2014relevance}
J.~Chen, Z.~Zhang, and Y.~Wang.
\newblock Relevance metric learning for person re-identification by exploiting
  global similarities.
\newblock In {\em ICPR}, pages 1657--1662, 2014.

\bibitem{cheng2011custom}
D.~S. Cheng, M.~Cristani, M.~Stoppa, L.~Bazzani, and V.~Murino.
\newblock Custom pictorial structures for re-identification.
\newblock In {\em BMVC}, pages 1--11, 2011.

\bibitem{chopra2005learning}
S.~Chopra, R.~Hadsell, and Y.~LeCun.
\newblock Learning a similarity metric discriminatively, with application to
  face verification.
\newblock In {\em CVPR}, pages 539--546, 2005.

\bibitem{ciresan2011flexible}
D.~C. Ciresan, U.~Meier, J.~Masci, L.~Maria~Gambardella, and J.~Schmidhuber.
\newblock Flexible, high performance convolutional neural networks for image
  classification.
\newblock In {\em IJCAI}, pages 1237--1242, 2011.

\bibitem{davis2007information}
J.~V. Davis, B.~Kulis, P.~Jain, S.~Sra, and I.~S. Dhillon.
\newblock Information-theoretic metric learning.
\newblock In {\em ICML}, pages 209--216, 2007.

\bibitem{farenzena2010person}
M.~Farenzena, L.~Bazzani, A.~Perina, V.~Murino, and M.~Cristani.
\newblock Person re-identification by symmetry-driven accumulation of local
  features.
\newblock In {\em CVPR}, pages 2360--2367, 2010.

\bibitem{fisher1936use}
R.~A. Fisher.
\newblock The use of multiple measurements in taxonomic problems.
\newblock {\em Annals of eugenics}, pages 179--188, 1936.

\bibitem{gheissari2006person}
N.~Gheissari, T.~B. Sebastian, and R.~Hartley.
\newblock Person reidentification using spatiotemporal appearance.
\newblock In {\em CVPR}, pages 1528--1535, 2006.

\bibitem{goldberger2004neighbourhood}
J.~Goldberger, G.~E. Hinton, S.~T. Roweis, and R.~Salakhutdinov.
\newblock Neighbourhood components analysis.
\newblock In {\em NIPS}, pages 513--520, 2004.

\bibitem{gong2014person}
S.~Gong, M.~Cristani, S.~Yan, and C.~C. Loy.
\newblock {\em Person re-identification}.
\newblock 2014.

\bibitem{gray2007evaluating}
D.~Gray, S.~Brennan, and H.~Tao.
\newblock Evaluating appearance models for recognition, reacquisition, and
  tracking.
\newblock In {\em PETS}, 2007.

\bibitem{gray2008viewpoint}
D.~Gray and H.~Tao.
\newblock Viewpoint invariant pedestrian recognition with an ensemble of
  localized features.
\newblock In {\em ECCV}, pages 262--275. 2008.

\bibitem{niyogi2004locality}
X.~He and P.~Niyogi.
\newblock Locality preserving projections.
\newblock In {\em NIPS}, pages 153--160, 2003.

\bibitem{hirzer2012person}
M.~Hirzer, P.~M. Roth, and H.~Bischof.
\newblock Person re-identification by efficient impostor-based metric learning.
\newblock In {\em AVSS}, pages 203--208, 2012.

\bibitem{hirzer2012relaxed}
M.~Hirzer, P.~M. Roth, M.~K{\"o}stinger, and H.~Bischof.
\newblock Relaxed pairwise learned metric for person re-identification.
\newblock In {\em ECCV}, pages 780--793. 2012.

\bibitem{hu2014discriminative}
J.~Hu, J.~Lu, and Y.-P. Tan.
\newblock Discriminative deep metric learning for face verification in the
  wild.
\newblock In {\em CVPR}, pages 1875--1882, 2014.

\bibitem{koestinger2012large}
M.~Koestinger, M.~Hirzer, P.~Wohlhart, P.~M. Roth, and H.~Bischof.
\newblock Large scale metric learning from equivalence constraints.
\newblock In {\em CVPR}, pages 2288--2295, 2012.

\bibitem{kruskal1978multidimensional}
J.~B. Kruskal and M.~Wish.
\newblock {\em Multidimensional scaling}.
\newblock 1978.

\bibitem{kviatkovsky2013color}
I.~Kviatkovsky, A.~Adam, and E.~Rivlin.
\newblock Color invariants for person reidentification.
\newblock {\em TPAMI}, pages 1622--1634, 2013.

\bibitem{li2012human}
W.~Li, R.~Zhao, and X.~Wang.
\newblock Human reidentification with transferred metric learning.
\newblock In {\em ACCV}, pages 31--44, 2012.

\bibitem{li2014deepreid}
W.~Li, R.~Zhao, T.~Xiao, and X.~Wang.
\newblock Deepreid: Deep filter pairing neural network for person
  re-identification.
\newblock In {\em CVPR}, pages 152--159, 2014.

\bibitem{li2013learning}
Z.~Li, S.~Chang, F.~Liang, T.~S. Huang, L.~Cao, and J.~R. Smith.
\newblock Learning locally-adaptive decision functions for person verification.
\newblock In {\em CVPR}, pages 3610--3617, 2013.

\bibitem{liao2015person}
S.~Liao, Y.~Hu, X.~Zhu, and S.~Z. Li.
\newblock Person re-identification by local maximal occurrence representation
  and metric learning.
\newblock In {\em CVPR}, pages 2197--2206, 2015.

\bibitem{liao2010modeling}
S.~Liao, G.~Zhao, V.~Kellokumpu, M.~Pietik{\"a}inen, and S.~Z. Li.
\newblock Modeling pixel process with scale invariant local patterns for
  background subtraction in complex scenes.
\newblock In {\em CVPR}, pages 1301--1306, 2010.

\bibitem{liu2012person}
C.~Liu, S.~Gong, C.~C. Loy, and X.~Lin.
\newblock Person re-identification: What features are important?
\newblock In {\em ECCV Workshop}, pages 391--401, 2012.

\bibitem{loy2013person}
C.~C. Loy, C.~Liu, and S.~Gong.
\newblock Person re-identification by manifold ranking.
\newblock In {\em ICIP}, pages 3567--3571, 2013.

\bibitem{loy2009multi}
C.~C. Loy, T.~Xiang, and S.~Gong.
\newblock Multi-camera activity correlation analysis.
\newblock In {\em CVPR}, pages 1988--1995, 2009.

\bibitem{ma2013domain}
A.~J. Ma, P.~C. Yuen, and J.~Li.
\newblock Domain transfer support vector ranking for person re-identification
  without target camera label information.
\newblock In {\em ICCV}, pages 3567--3574, 2013.

\bibitem{ma2012local}
B.~Ma, Y.~Su, and F.~Jurie.
\newblock Local descriptors encoded by fisher vectors for person
  re-identification.
\newblock In {\em ECCV Workshop}, pages 413--422, 2012.

\bibitem{ma2014covariance}
B.~Ma, Y.~Su, and F.~Jurie.
\newblock Covariance descriptor based on bio-inspired features for person
  re-identification and face verification.
\newblock {\em ICIVC}, pages 379--390, 2014.

\bibitem{ma2014person}
L.~Ma, X.~Yang, and D.~Tao.
\newblock Person re-identification over camera networks using multi-task
  distance metric learning.
\newblock {\em TIP}, pages 3656--3670, 2014.

\bibitem{mignon2012pcca}
A.~Mignon and F.~Jurie.
\newblock Pcca: A new approach for distance learning from sparse pairwise
  constraints.
\newblock In {\em CVPR}, pages 2666--2672, 2012.

\bibitem{paisitkriangkrai2015learning}
S.~Paisitkriangkrai, C.~Shen, and A.~v.~d. Hengel.
\newblock Learning to rank in person re-identification with metric ensembles.
\newblock pages 1846--1855, 2015.

\bibitem{pedagadi2013local}
S.~Pedagadi, J.~Orwell, S.~Velastin, and B.~Boghossian.
\newblock Local fisher discriminant analysis for pedestrian re-identification.
\newblock In {\em CVPR}, pages 3318--3325, 2013.

\bibitem{prosser2010person}
B.~Prosser, W.-S. Zheng, S.~Gong, T.~Xiang, and Q.~Mary.
\newblock Person re-identification by support vector ranking.
\newblock In {\em BMVC}, pages 1--11, 2010.

\bibitem{roweis2000nonlinear}
S.~T. Roweis and L.~K. Saul.
\newblock Nonlinear dimensionality reduction by locally linear embedding.
\newblock {\em Science}, pages 2323--2326, 2000.

\bibitem{tao2013person}
D.~Tao, L.~Jin, Y.~Wang, Y.~Yuan, and X.~Li.
\newblock Person re-identification by regularized smoothing kiss metric
  learning.
\newblock {\em TCSVT}, pages 1675--1685, 2013.

\bibitem{vezzani2013people}
R.~Vezzani, D.~Baltieri, and R.~Cucchiara.
\newblock People reidentification in surveillance and forensics: A survey.
\newblock {\em ACM Computing Surveys (CSUR)}, page~29, 2013.

\bibitem{wang2007shape}
X.~Wang, G.~Doretto, T.~Sebastian, J.~Rittscher, and P.~Tu.
\newblock Shape and appearance context modeling.
\newblock In {\em ICCV}, pages 1--8, 2007.

\bibitem{weinberger2005distance}
K.~Q. Weinberger, J.~Blitzer, and L.~K. Saul.
\newblock Distance metric learning for large margin nearest neighbor
  classification.
\newblock In {\em NIPS}, pages 1473--1480, 2005.

\bibitem{wold1987principal}
S.~Wold, K.~Esbensen, and P.~Geladi.
\newblock Principal component analysis.
\newblock {\em Chemometrics and intelligent laboratory systems}, pages 37--52,
  1987.

\bibitem{xiong2014person}
F.~Xiong, M.~Gou, O.~Camps, and M.~Sznaier.
\newblock Person re-identification using kernel-based metric learning methods.
\newblock In {\em ECCV}, pages 1--16. 2014.

\bibitem{yi2014deep}
D.~Yi, Z.~Lei, and S.~Z. Li.
\newblock Deep metric learning for practical person re-identification.
\newblock {\em arXiv preprint arXiv:1407.4979}, 2014.

\bibitem{zhao2013person}
R.~Zhao, W.~Ouyang, and X.~Wang.
\newblock Person re-identification by salience matching.
\newblock In {\em ICCV}, pages 2528--2535, 2013.

\bibitem{zhao2013unsupervised}
R.~Zhao, W.~Ouyang, and X.~Wang.
\newblock Unsupervised salience learning for person re-identification.
\newblock In {\em CVPR}, pages 3586--3593, 2013.

\bibitem{zhao2014learning}
R.~Zhao, W.~Ouyang, and X.~Wang.
\newblock Learning mid-level filters for person re-identification.
\newblock In {\em CVPR}, pages 144--151, 2014.

\bibitem{zheng2013reidentification}
W.-S. Zheng, S.~Gong, and T.~Xiang.
\newblock Reidentification by relative distance comparison.
\newblock {\em TPAMI}, pages 653--668, 2013.

\end{thebibliography}
}
\end{document}